\documentclass[10pt,twocolumn,letterpaper]{article}

\usepackage{iccv}              
\usepackage{soul}
\usepackage{float}
\usepackage{cuted}
\usepackage{multirow}

\definecolor{iccvblue}{rgb}{0.21,0.49,0.74}

\usepackage[pagebackref,breaklinks,colorlinks,allcolors=iccvblue]{hyperref}
\usepackage{tabularx}
\usepackage{booktabs}
\title{SpeakerVid-5M: A Large-Scale High-Quality Dataset for Audio-Visual Dyadic Interactive Human Generation}
\def\paperID{8888} 
\def\confName{ICCV}
\def\confYear{2025}

\author{Youliang Zhang$^{1,2}$
    \hspace{0.15in}
    Zhaoyang Li$^{2}$
    \hspace{0.15in}
    Duomin Wang$^{2}$\textsuperscript{\textdagger}
    \hspace{0.15in}
    Jiahe Zhang\\
    Deyu Zhou$^{2, 3}$
    \hspace{0.15in}
    Zixin Yin$^{2, 4}$
    \hspace{0.15in}
    Xili Dai$^{3}$
    \hspace{0.15in}
    Gang Yu$^{2}$
    \hspace{0.15in}
    Xiu Li$^{1}$\textsuperscript{\textdaggerdbl}\\
    $^1$Tsinghua University.\hspace{0.15in} $^2$StepFun.\\
    $^3$The Hong Kong University of Science and Technology (Guangzhou).\\
    $^4$The Hong Kong University of Science and Technology.\\
     {\tt\small zhangyou24@mails.tsinghua.edu.cn, lizhaoyang@mail.ustc.edu.cn}\\
     {\tt\small (wangduomin,daixili.cs, jiaahe0111)@gmail.com}\\
      {\tt\small zyinaf@connect.ust.hk, dzhou861@connect.hkust-gz.edu.cn}\\
      {\tt\small yugang@stepfun.com, li.xiu@sz.tsinghua.edu.cn}
    }

\begin{document}

\maketitle

\begingroup\def\thefootnote{\textdagger}\footnotetext{Project Lead}\endgroup
\begingroup\def\thefootnote{\textdaggerdbl}\footnotetext{Corresponding author}\endgroup

\begin{strip}
\centering
    \vspace{-50pt}
    \includegraphics[width=\textwidth]{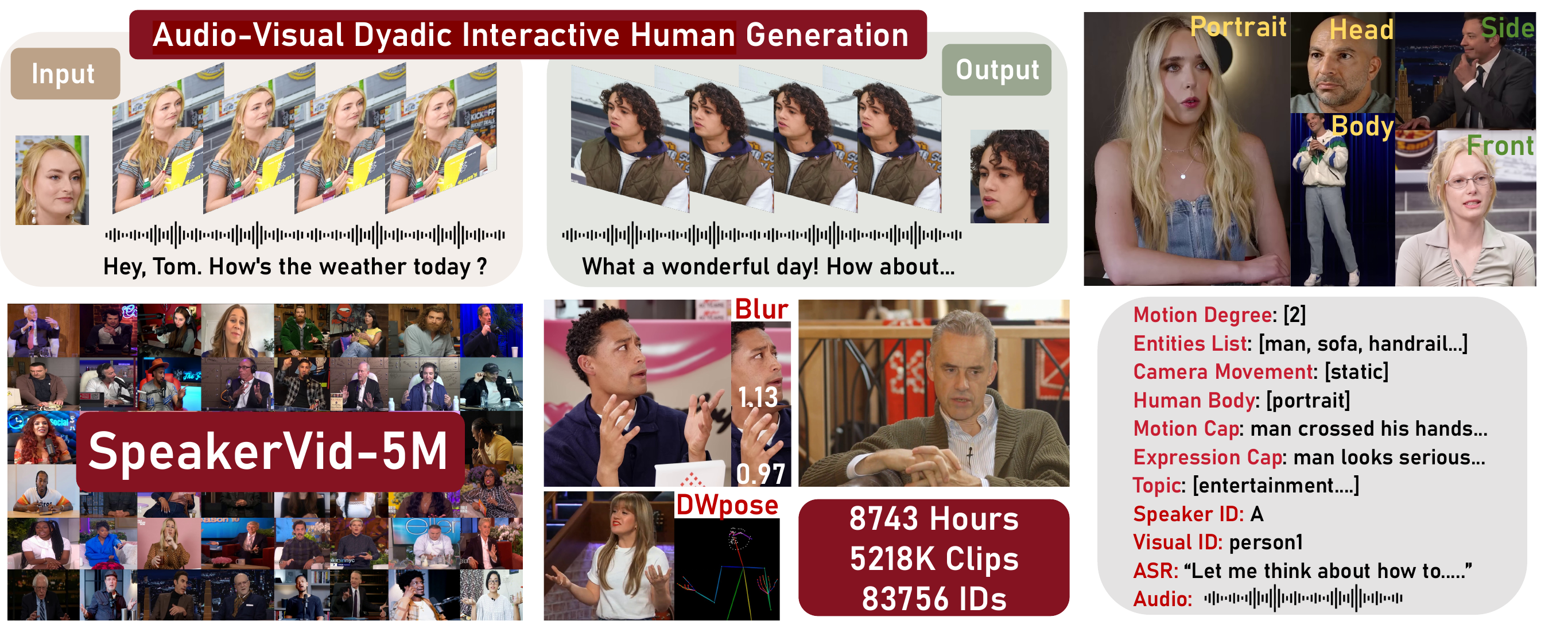}
    \captionof{figure}{
    {\textbf{Overview of the audio-visual dyadic generation task and the SpeakerVid-5M dataset.} The primary task (top row) is to generate a coherent audio-visual response based on the input of initiator. Our SpeakerVid-5M (bottom left) provides over 8.7K hours data to facilitate this research. Each clip is enriched with detailed multi-modal annotations (right panel), enabling fine-grained generation.} }
    \label{fig:motivation}
\end{strip}

\vspace{-20pt}
\begin{abstract}

The rapid development of large-scale models has catalyzed significant breakthroughs in the digital human domain. These advanced methodologies offer high-fidelity solutions for avatar driving and rendering, leading academia to focus on the next major challenge: audio-visual dyadic interactive virtual human. To facilitate research in this emerging area, we present SpeakerVid-5M dataset, the first large-scale, high-quality dataset designed for audio-visual dyadic interactive virtual human generation. Totaling over $8,743$ hours, SpeakerVid-5M contains more than $5.2$ million video clips of human portraits. It covers diverse scales and interaction types, including monadic talking, listening, and dyadic conversations. Crucially, the dataset is structured along two key dimensions: interaction type and data quality. First, it is categorized into four types(dialogue branch, single branch, listening branch and multi-turn branch) based on the interaction scenario. Second, it is stratified into a large-scale pre-training subset and a curated, high-quality subset for Supervised Fine-Tuning (SFT). This dual structure accommodates a wide array of 2D virtual human tasks. In addition, we provide an autoregressive (AR)-based video chat baseline trained on this data, accompanied by a dedicated set of metrics and test data to serve as a benchmark (\textbf{VidChatBench}) for future work. Both the dataset and the corresponding data processing code will be publicly released. Project page: \href{https://dorniwang.github.io/SpeakerVid-5M/}{https://dorniwang.github.io/SpeakerVid-5M/}.
\end{abstract}

\section{Introduction}
\label{sec:intro}

In the era of rapid advancements in large-scale models, large video models have acquired the capability to generate high-fidelity video, presenting unprecedented advantages for the generation and driving of high-quality virtual humans. Pioneering works based on Generative Adversarial Networks (GANs)~\cite{zhou2021pose,wang2023progressive,yu2022talking}, established the foundation for realistic 2D virtual human driving and rendering. However, the advent of large video models exemplified by diffusion-based works~\cite{lin2025omnihuman,chen2025hunyuanvideo,wang2025fantasytalking,luo2025dreamactor,wei2025mocha} has achieved state-of-the-art realism, significantly elevating the authenticity of both driving and rendering.

This leap in fidelity has enabled broader industrial adoption, facilitating commercial-grade applications in areas such as automated lip-syncing, digital newscasting, and virtual actors. Nevertheless, a more ambitious goal has captured the attention of researchers in both academia and industry: the creation of proactive, interactive virtual humans. This endeavor is akin to equipping an avatar with a ``brain'', advancing towards virtual beings that are not just passively driven but can engage autonomously. Such technology holds compelling potential for applications like more realistic virtual assistants, live-streaming e-commerce, and online education. Several works have approached this task from a system-building perspective~\cite{wang2025agentavatar,zhu2025infp,qi2025chatanyone,low2025talkingmachines}, have constructed functional interactive agents by integrating existing, off-the-shelf modules. In contrast to earlier systems that lack the ability for multimodal perception and understanding, BodyofHer~\cite{ao2024body} presented an end-to-end trained interactive agent based on autoregressive LLM paradigm that accepts a full suite of multimodal inputs. These works collectively represent pioneering explorations in this direction.

In the current research paradigm, which is increasingly centered on foundation models, the development of task-specific applications necessitates large-scale training datasets. However, a notable scarcity of open-source datasets focused on interactive virtual humans persists within the academic community. 


Training foundation models for interactive virtual humans requires vast amounts of specialized data, yet such resources are scarce. To address this critical need, we introduce \textbf{SpeakerVid-5M}, the first large-scale, high-quality dataset for audio-visual dyadic interaction, featuring richly annotated and meticulously aligned audiovisual pairs.

SpeakerVid-5M features:
(1) \textbf{Large-scale data}:
The dataset contains 770K high-quality dynadic conversion audiovisual pairs (totaling 1.8K hours) and 5.2M high-quality single-speaker clips (totaling 8.7K hours).
(2) \textbf{High resolution}:
93\% of the videos are in 1080P or higher, ensuring detailed and clear visual input for generation tasks.
(3) \textbf{Rich annotation types}:
Each clip is accompanied by structured textual annotations, skeletal sequences, ASR transcriptions, and blur scores, supporting a wide range of multimodal learning objectives.
(4) \textbf{High-quality}:
Precise synchronization between audio and video is ensured, with rigorous filtering applied to both modalities to guarantee clean and reliable training data. 
(5) \textbf{Body compositional diversity}: Data instance is captured with labels spanning full-body, half-body, head-only, and side-view profiles, enabling fine-grained control over framing analysis.
In addition, we select $500$ videos from out-of-distribution speaker IDs to construct the \textbf{VidChatBench} benchmark for the audio-visual dyadic interactive virtual human task, which focuses on video quality, audio-visual consistency, dialogue coherence, and identity preservation for 
model performance evaluation.

SpeakerVid-5M is designed to facilitate research in interactive virtual human task. The rich annotations also make it a valuable resource for a variety of related tasks, such as human animation, talking head generation, multimodal dialogue modeling, etc.\, all of which face the challenge of a lack of large-scale, high-quality datasets.
In addition, we conduct an initial exploration of implementing audio-visual dyadic interactive virtual human generation under an autoregressive paradigm. Given input video and audio, the model jointly generates the speaker’s response in both audio and visual modalities.

Our contributions can be summarized as follows:
\begin{itemize}
    \item We propose SpeakerVid-5M, the first large-scale dataset designed specifically for the audio-visual dyadic interactive virtual human task. It includes 1M high-quality dialogue audiovisual pairs, with supporting for multi-turn conversations. The VidChatBench is also provided for better evaluation.

    \item SpeakerVid-5M contains 5M single-speaker audiovisual clips, making it the largest talking human dataset. It covers a wide range of annotated visual formats, including talking heads, half-body, full-body, and side-view videos.
    
    \item We open-source the entire dataset, including the raw data, annotations, and data processing pipeline, providing full transparency and reproducibility for the community.
\end{itemize} 

\section{Related Work} 
\label{sec:related}

\subsection{Audio-Visual Human Video Generation.}
Lip-sync~\cite{prajwal2020lip,li2024latentsync,peng2025omnisync} and talking head generation~\cite{zhou2021pose,wang2023progressive,tian2024emo,yu2022talking,xu2024vasa} are foundational tasks in audio-driven human video generation, among those, PD-FGC~\cite{wang2023progressive} was a pioneering work that first achieved the generation of talking heads with vivid facial expressions. Additionally, tasks such as learning to listen~\cite{ng2023can,ng2022learning} also fall under the category of audio-driven portrait generation, focusing on responsive non-verbal behaviors. Subsequently, with the rapid development of foundational video models, the field has bifurcated into two primary research directions. One branch has pushed for higher quality and broader scope~\cite{tian2024emo,wei2025mocha,luo2025dreamactor,wang2025fantasytalking,chen2025hunyuanvideo,lin2025omnihuman}, with EMO~\cite{tian2024emo} introducing diffusion video models to achieve state-of-the-art realism, and works like OmniHuman-1~\cite{lin2025omnihuman}, MoCha~\cite{wei2025mocha}, and Veo3~\cite{veo3} expanding generation to full-body, multi-agent performances, and direct text-to-audiovisual synthesis. The second branch focuses on creating interactive virtual humans~\cite{wang2025agentavatar,zhu2025infp,qi2025chatanyone,low2025talkingmachines,ao2024body}. This includes modular systems like AgentAvatar~\cite{wang2025agentavatar} that integrate LLMs as planners, and end-to-end paradigms like BodyofHer~\cite{ao2024body} that enable direct multimodal understanding for autonomous reactions. These pioneering efforts in interactivity shift the focus from pure driving alignment to agent autonomy.

\subsection{Audio-Visual Human Video Datasets.}
Early research in talking head and lip-sync generation initially leveraged datasets from related fields such as lip reading~\cite{afouras2018lrs3,afouras2018deep,chung2017lip} and speaker recognition~\cite{nagrani2017voxceleb,chung2018voxceleb2}, Subsequently, datasets specifically curated for these tasks began to emerge~\cite{yu2023celebv,zhang2021flow,wang2021one}. ViCo~\cite{zhou2022responsive} introduced a dataset for the ``learning to listen'' task, though it was focused on monadic scenarios. However, these early datasets were often limited in scale and of lower quality, rendering them inadequate for the demands of current high-quality, data-intensive models. While massive datasets exist, they are either too general-purpose and of variable quality (\textit{e.g.}, ACAV-100M~\cite{lee2021acav100m}) or proprietary and unreleased (\textit{e.g.}, the data used by OmniHuman-1~\cite{lin2025omnihuman}). The recent release of OpenHumanVid~\cite{li2025openhumanvid} provided a valuable resource, but it was only partially released and remains focused on monadic, talking-head scenarios. This leaves a critical void for a public, large-scale dataset centered on audio-visual dyadic interactions for virtual human research. Our work, SpeakerVid-5M, is designed to fill this void.

\section{Dataset Curation}

\begin{figure*}[t]
    \centering
\includegraphics[width=\textwidth]{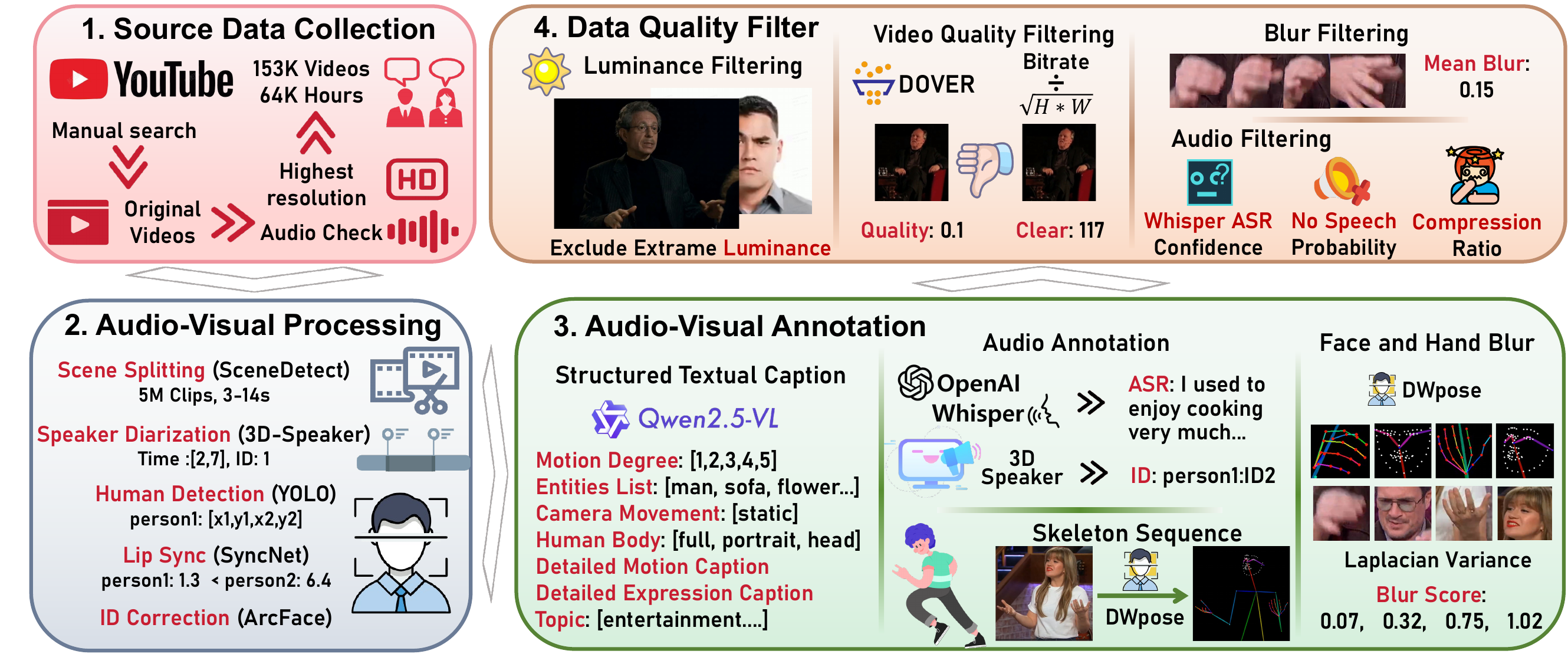}
    \caption{\label{fig:pipeline}
    {\textbf{The SpeakerVid-5M curation pipeline.} The process consists: (1) Source data collection from YouTube; (2) Multi-step audio-visual pre-processing; (3) Rich multi-modal annotation using models like Qwen-VL; (4) Rigorous quality filtering stage for data fidelity.}
    }
\end{figure*}

The construction process of our SpeakerVid-5M dataset mainly includes four parts: Source data collection, Data pre-processing, Data annotation, and Quality filter. The entire construction process is illustrated in Figure \ref{fig:pipeline}.
\subsection{Source Data Collection}
We manually collected high-quality videos featuring two-person dialogue from YouTube. The sources mainly include interviews, news reports, seminars, television programs, variety shows, debates, and educational videos. The collected videos exhibit diverse formats such as full-body, half-body, frontal, and side views, all with clear facial visibility and corresponding audio. In total, we gathered $153$K audio-visual videos, amounting to $64,386$ hours of raw data, with $93\%$ of the videos in 1080P resolution or higher. The temporal scope of the dataset extends from $2006$ to the present day (June $2025$). 
This collection spans a diverse range of genres, mainly including entertainment, people and blogs, comedy, news and politics, education, science, and sports.

\subsection{Data Pre-processing}
\textbf{Scene splitting}. We employed SceneDetect \cite{scenedetect2024} for scene splitting, which identifies visually significant transitions by analyzing changes in color and brightness to determine scene boundaries. Based on the initial segmentation results, we conducted further post-processing by discarding clips shorter than $3$ seconds and splitting those longer than $14$ seconds. As a result, we obtained video clips ranging from $3$ to $14$ seconds in length. We record the temporal order and the human ID of each segment, allowing clips to be easily concatenated to form longer sequences when needed. The segments resulting from this stage are denoted by \textit{$S_{sp}$}.


\textbf{Speaker diarization}. We utilize 3D-Speaker \cite{chen20243d} to perform speaker diarization, segmenting the original audio (extracted directly from video before scene splitting) into multiple segments and recording the timestamps and corresponding speaker IDs for each segment. For each original audio, we identified the two primary speaker IDs based on the frequency and duration of their speech. Segments associated with other speaker IDs were discarded in the subsequent processing steps. The segments resulting from this stage are denoted by \textit{$S_{sv}$}.

\textbf{Human detection}. We apply YOLO \cite{yolov8} for human tracking within each video clip, and aggregate the tracking results over time to form a single spatiotemporal visual track for each individual. Because the scene boundaries provided by SceneDetect can be inaccurate, we further refine each clip by temporally and spatially cropping it based on YOLO detection results to obtain usable single-speaker video clips. This stage may yield multiple clips, each originating from different spatial regions within the same temporal window of a single video clip \textit{$S_{sp}$}. The resulting segments from this stage are denoted by \textit{$S_{rsp}$}.

\textbf{Lip synchronization}. For each video clip obtained from the last stage, we first calculate the temporal overlap between \textit{$S_{rsp}$} and \textit{$S_{sv}$}, this yields the overlapped segment \textit{$S_{ol}$}. We then employ SyncNet \cite{raina2022syncnet} to perform audio-visual synchronization within each segment \textit{$S_{ol}$}. The SyncNet confidence score is then used to associate each speaker ID with a specific visual bounding box detected by YOLO. In cases where two individuals are present and conversing within the same temporal window, we assign speaker ID to the bounding box with the highest confidence score.

\textbf{ID correction.}
To further verify and refine the speaker IDs obtained through the collaboration of Speaker diarization and lip-sync analysis, we employ a vision-based method, ArcFace \cite{deng2019arcface}, for further correction. For multiple clips extracted from the same original video, the speaker IDs derived from audio segmentation should be consistent with those inferred from visual information.
We first assign speaker IDs based on audio segmentation, then compute facial cosine similarity across clips belonging to the same speaker ID. Outliers identified via low similarity scores are compared against other speaker IDs using ArcFace. If a higher similarity is found with another ID, the outliers' ID is updated accordingly, completing the correction process.

\subsection{Data Annotation}
\textbf{Structured Textual Caption.}
Textual captions are provided in detailed structure formats to ensure a comprehensive representation of the video content.  
We leverage the powerful multimodal model Qwen2.5-VL \cite{bai2025qwen2} to generate textual annotations for our video data, which demonstrates remarkable capabilities in visual understanding and instruction following. 
The structured annotations include camera movement patterns, a list of entities present in the video, body orientation (front or side view), and whether the person is shown in a half-body or full-body. Additionally, we provide detailed descriptions of human body movements and facial expressions.
Furthermore, we use Qwen-3 \cite{yang2025qwen3} to summarize automatic speech recognition results across multiple clips originating from the same source video. Based on these summaries, we annotate the topic category of the two-person dialogue presented in the video.

\textbf{Audio Annotation.}
Each data sample in our dataset consists of a speaker video with well-aligned audio and visual modalities. To support tasks related to audio generation and control, we annotate the audio with several key features. First, we apply Whisper \cite{radford2023robust} for automatic speech recognition to obtain text transcriptions of the audio. Second, we annotate each audio segment with its corresponding SyncNet metrics, including the audio-visual offset, synchronization confidence score, and audio-visual embedding distance. Finally, we use 3D-Speaker to assign speaker IDs to multiple audio segments from a single original audio. 
For each clip, in addition to the original audio, we provide a cleaned version in which segments not belonging to the target speaker ID are replaced with silent audio.

\textbf{Skeleton Sequence.}
Building upon the spatiotemporal human bounding boxes obtained from YOLO, we further utilize DWpose \cite{yang2023effective} for human pose estimation. The resulting skeletal sequences include keypoints of the face, hands, and body. Clips without a detected face are filtered out.

\textbf{Face and Hand Blur Score.}
We extract face and hand bounding boxes from the DWpose annotations. For each frame, we crop the pixel regions corresponding to the face and both hands, then resize these crops to a resolution of $128\times128$. Next, we calculate the Laplacian variance of the cropped face and hand images using the Laplacian operator. A higher variance typically indicates a sharper, clearer image, which we use as a clarity score. Since videos containing limb movements often exhibit motion blur during rapid movements, incorporating this clarity score as a conditioning factor may enhance the model’s performance in handling such scenarios \cite{lin2025cyberhost}.

\textbf{Motion score.}
The range of human motion plays a crucial role in generating human-centered videos. Compared to actions in general videos, human movements are more subjective and varied. Therefore, we employed Qwen2.5-VL to simulate human judgment and score the motion magnitude of individuals in the videos. To better simulate the behavior of different annotators, we employ multiple prompts, each embodying a distinct persona, to rate each video. The ratings are on a $1$-to-$5$ scale, where a score of $1$ indicates minimal movement and a score of $5$ denotes a high motion amplitude. After excluding outliers, the average score is taken as the final motion magnitude annotation for the video.

\subsection{Quality Filter}
\textbf{Luminance Filtering.}
Following OpenHumanVid \cite{li2025openhumanvid}, we calculate the luminance score to filter overly dark or bright videos. The luminance score is calculated using the formula: $0.2126 × R + 0.7152 × G + 0.0722 × B$, where $R$, $G$, and $B$ denote the pixel values of the red, green, and blue channels, respectively. Videos with a luminance score below $10$ or above $210$ are filtered out.

\textbf{Video Quality Filtering.}
To ensure the visual quality of the video data, we employed the DOVER \cite{wu2023exploring} model for video quality assessment. DOVER decomposes each video into aesthetics-related and technical-related components and evaluates them separately. This approach enhances the consistency between the model’s assessment results and human subjective perception of video quality. We filtering out videos with fused scores below $0.25$.

\textbf{Clear score Filtering.}
Resolution alone is insufficient to assess the visual clarity of a video, as some high-resolution videos may still exhibit blurriness and other low-quality characteristics. Bitrate reflects the amount of information carried per frame and provides an informative quality measure. 
We compute a clarity score as ${B} / \sqrt{{W} \times {H}}$, where $W$ and $H$ means the resolution of a video and $B$ is the bitrate.
Videos with clarity scores in the bottom $5\%$ are filtered out.

\textbf{Blur Filtering.}
Human-centered videos often suffer from motion blur, especially when the subject exhibits large movements. In our dataset, such blur typically affects the face and hands. To address this, we compute the average blur score over face and hand for each frame in a video. Videos with an average blur score(face or hand) below $0.1$ are filtered out to ensure data quality.

\textbf{Audio Filtering.}
Previous audio-visual datasets have typically focused solely on visual quality while neglecting the quality of the audio component.
We additionally focus on four key metrics in the ASR process, confidence score, no-speech probability, compression ratio, and detected language. These indicators reflect the reliability of the ASR output, whether the audio contains human speech, and whether there are corrupted or non-speech signals in the audio. At this stage, we filter out audio clips with a confidence score(average log probability) lower than $-1.5$, a no-speech probability greater than $0.8$, or a compression ratio exceeding $2.5$. To further ensure the quality of the detected results, we filter out samples in which the detected language mismatches the language labeled in the video's metadata.
\section{Dataset Statistics and Analysis}

\begin{figure*}[htbp]
    \centering
    \begin{subfigure}[t]{0.32\textwidth}
        \centering
        \includegraphics[width=\linewidth]{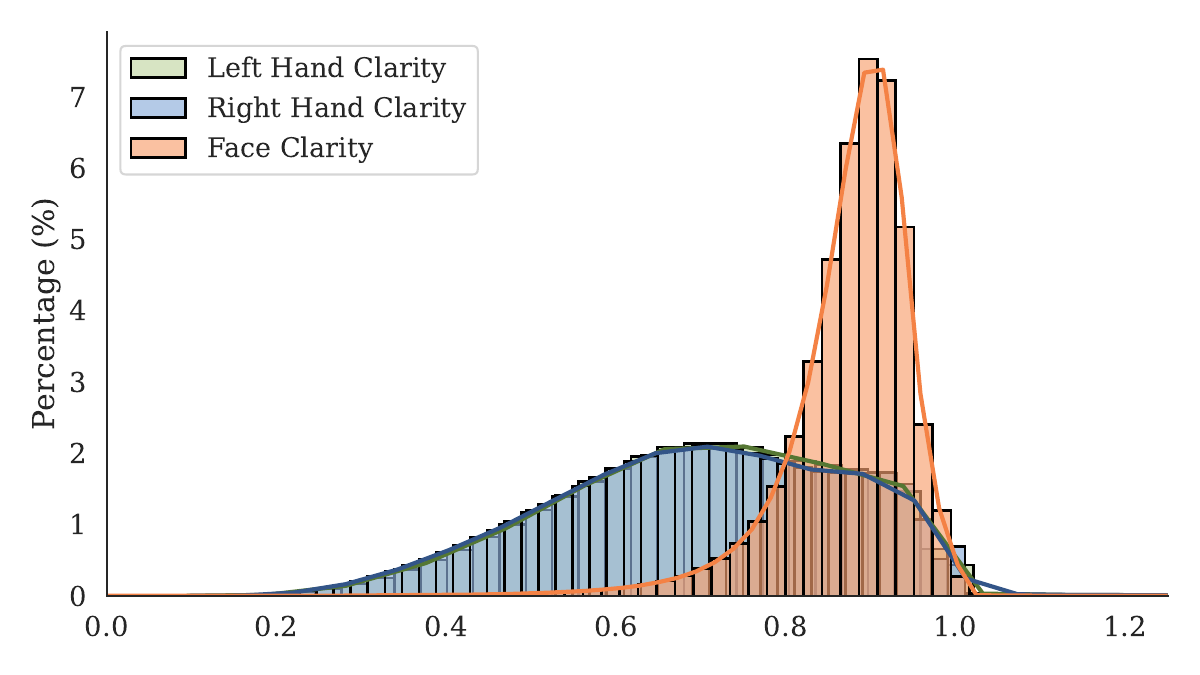}
        \caption{Face and Hand Blur Score}
        \label{fig:a}
    \end{subfigure}
    \hfill
    \begin{subfigure}[t]{0.2\textwidth}
        \centering
        \includegraphics[width=\linewidth]{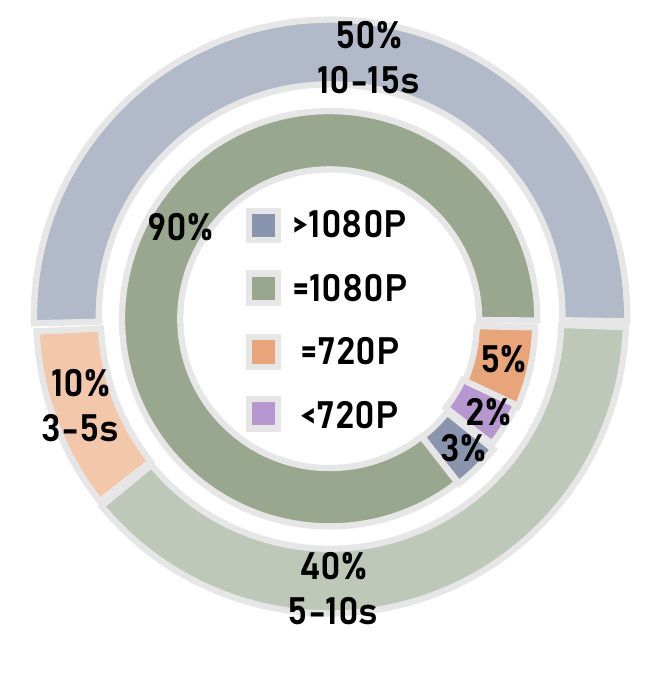}
        \caption{Duration and Resolution}
        \label{fig:b}
    \end{subfigure}
    \hfill
    \begin{subfigure}[t]{0.22\textwidth}
        \centering
        \includegraphics[width=\linewidth]{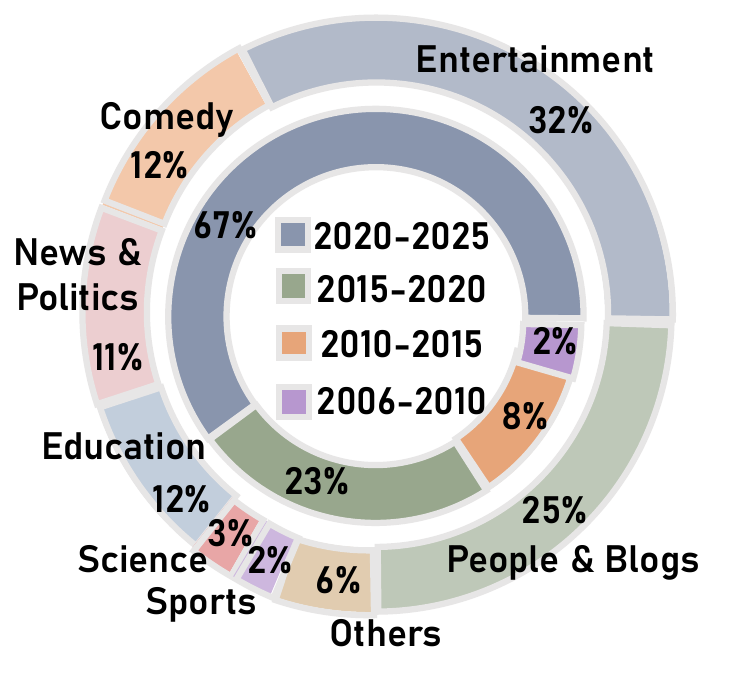}
        \caption{Topic and Year Distribution}
        \label{fig:c}
    \end{subfigure}
    \hfill
    \begin{subfigure}[t]{0.2\textwidth}
        \centering
        \includegraphics[width=\linewidth]{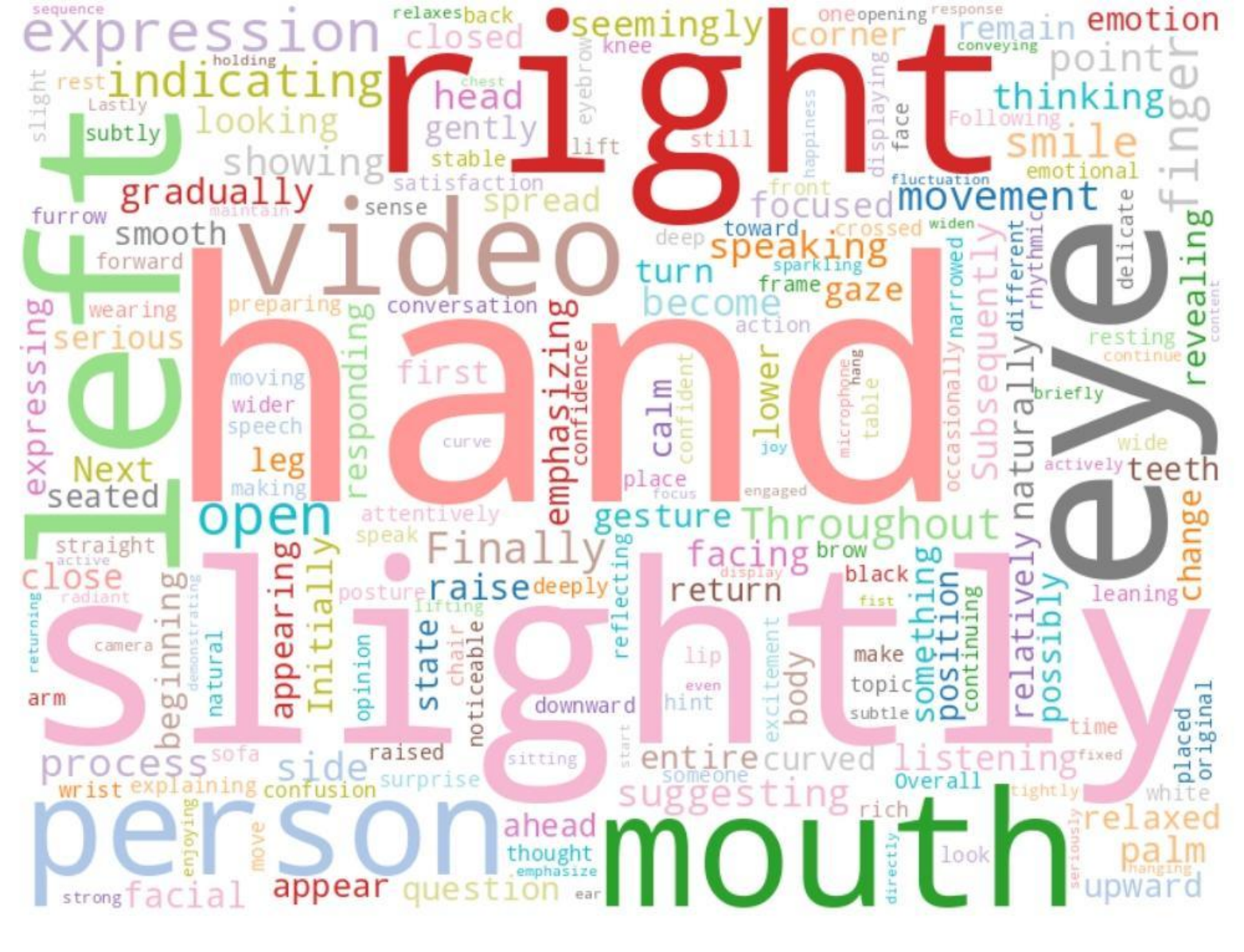}
        \caption{Caption Word Cloud}
        \label{fig:d}
    \end{subfigure}

    \vspace{1em} 

    \begin{subfigure}[t]{0.32\textwidth}
        \centering
        \includegraphics[width=\linewidth]{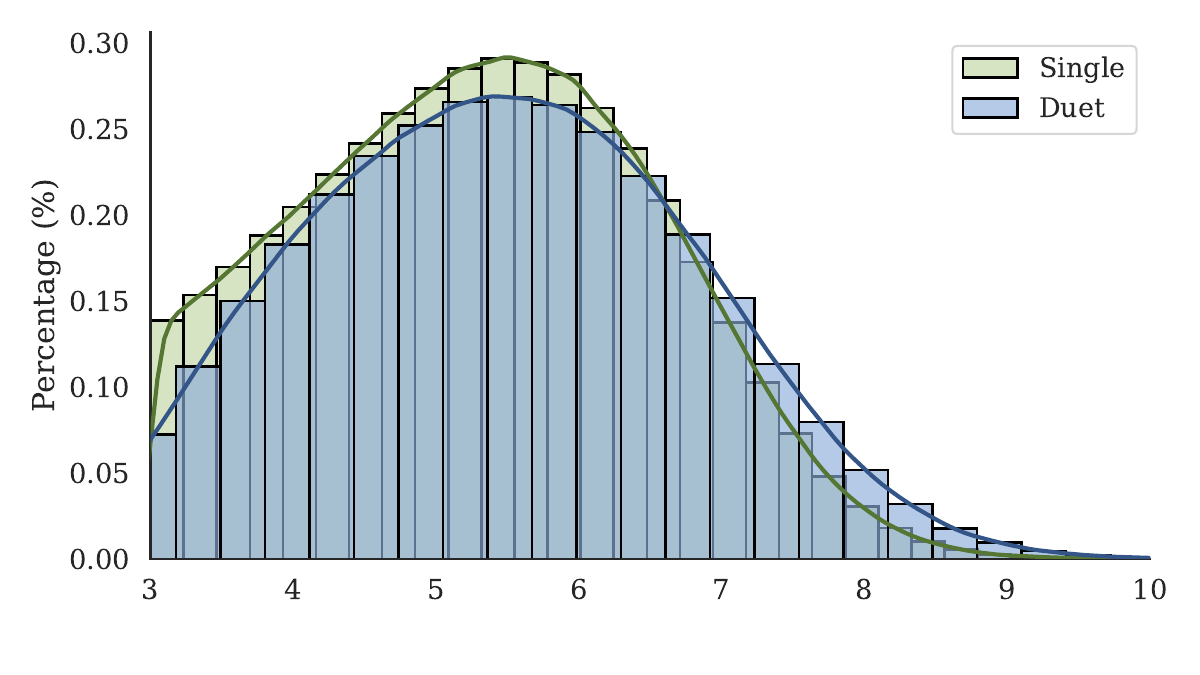}
        \caption{Sync Conf Score Distribution}
        \label{fig:e}
    \end{subfigure}
    \hfill
    \begin{subfigure}[t]{0.32\textwidth}
        \centering
        \includegraphics[width=\linewidth]{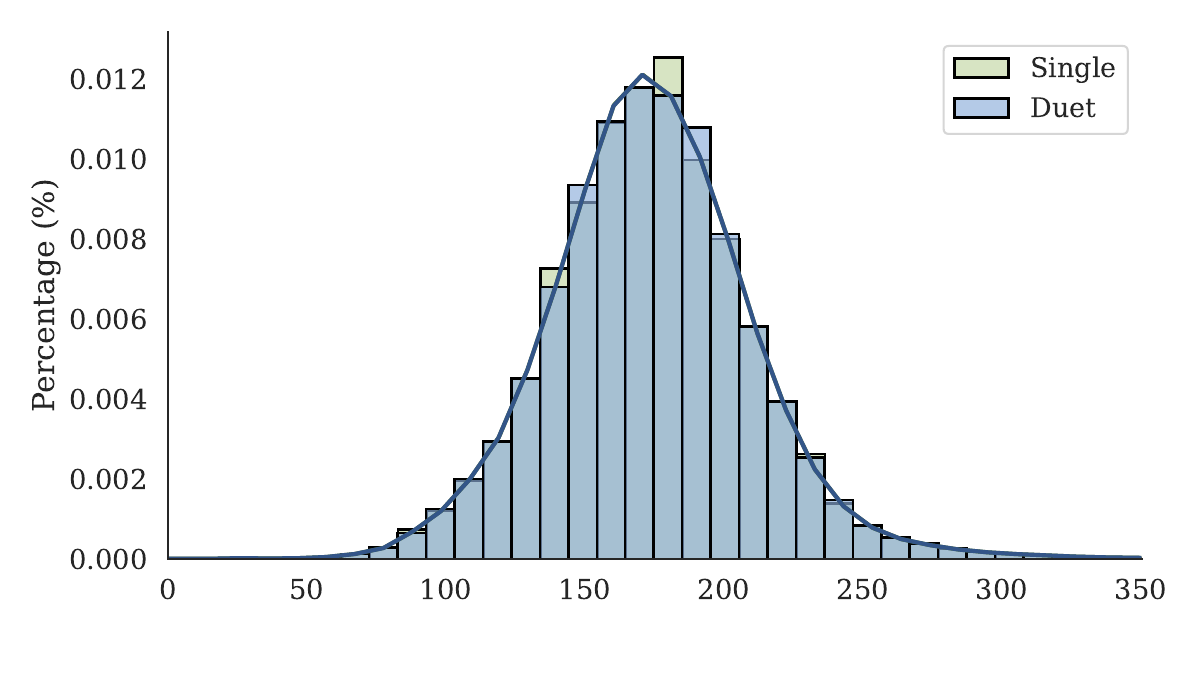}
        \caption{Caption Words Distribution}
        \label{fig:f}
    \end{subfigure}
    \hfill
    \begin{subfigure}[t]{0.32\textwidth}
        \centering
        \includegraphics[width=\linewidth]{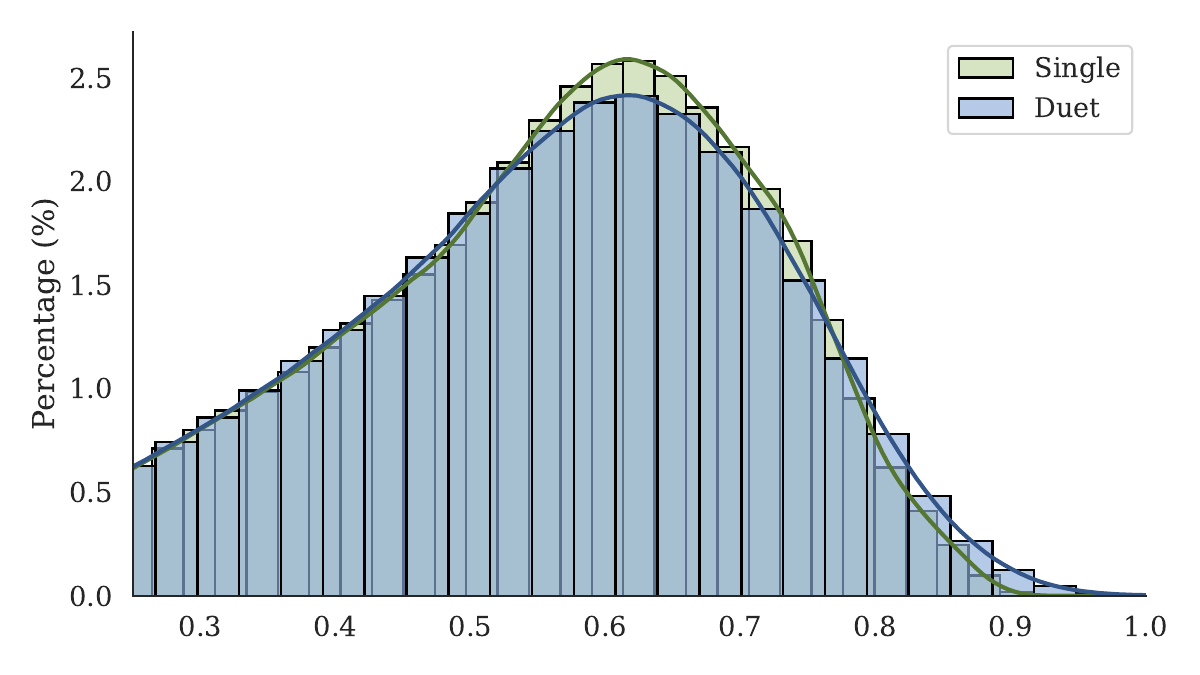}
        \caption{DOVER Quality Score Distribution}
        \label{fig:g}
    \end{subfigure}

    \caption{Statistics of our dataset from multiple aspects, including blur score, sync score, caption, etc.}
    \label{fig:dataset_stats}
\end{figure*}

\begin{table*}[!thb]
\begin{center}
\small
\resizebox{\textwidth}{!}
{
\begin{tabular}{lccccccccccccc}
\toprule
Datasets& Domain& Clips & Duration (hours) & Generation & Person num & audio & pose & Speaker ID & Blur anno & Body composition &  Caption type & IDs & Resolution\\
\midrule
UCF-101 \cite{soomro2012ucf101} & Human & 13.3K & 26.7 & --& N/A & & & & && Text & N/A & 240P \\
ActivityNet \cite{caba2015activitynet} & Human & 100K & 849 & -- & N/A & &  & & && Text & N/A & N/A \\
NTU RGB+D \cite{shahroudy2016ntu} & Human & 114K & 3.7 & Conditioned & single & & \checkmark & & & & - & N/A & 1080P \\
TikTok-v4 \cite{chang2023magicdance} & Human & 350 & 1 & Conditioned & single & & \checkmark & & & &  - & N/A & N/A \\
Openhumanvid \cite{li2025openhumanvid} & Human & 13.4M & 16.7K & Conditioned & multi & \checkmark & \checkmark & & & & Structured & N/A & 720P \\
\midrule
VoxCeleb \cite{nagrani2017voxceleb} & Head & 21.2K & 352 & Conditioned & single & \checkmark & & & & &  - & 1.2k & 224P \\
VoxCeleb2 \cite{chung2018voxceleb2} & Head & 150.4K & 2.4K & Conditioned & single & \checkmark & & & & &  - & 6.1K & 224P \\
MEAD \cite{wang2020mead} & Head & 281.4K & 39 & Conditioned & single & \checkmark & & & & &  - & 60 & 1080P \\
CelebV-HQ \cite{zhu2022celebv} & Head & 35.6K & 68 & Conditioned & single & \checkmark & & & & & Structured & 15.6K & 512P \\
CelebV-Text \cite{yu2023celebv} & Head & 70K & 279 & Conditioned & single & \checkmark & & & & & Structured & N/A & 512P \\
\midrule
SpeakerVid-5M & Human & 5.2M & 8.7K & Conditioned & single & \checkmark & \checkmark &\checkmark & \checkmark&\checkmark& Structured & 83K & 1080P \\
SpeakerVid-5M (Dialogue) & Human & 770K & 1.8K & Dyadic & single & \checkmark & \checkmark & \checkmark&\checkmark&\checkmark& Structured & 16K & 1080P \\
\bottomrule
\end{tabular}
}
\caption{\small{
{\textbf{Comparative analysis of SpeakerVid-5M with existing human video datasets.} Person num indicates the number of individuals present in a given clip. For each clip, only one person is retained, ensuring a clean alignment between the audio and visual streams. Blur anno represents the degree of blurriness of the hands and face in each frame. Body composition means the fine-grained annotations for body composition (full-body, half-body, head-only) and camera perspective (frontal, side), features that are absent in most prior work.}
}}
\label{tab:compare}
\end{center}
\end{table*}

\begin{figure*}[t]
    \centering
\includegraphics[width=\textwidth]{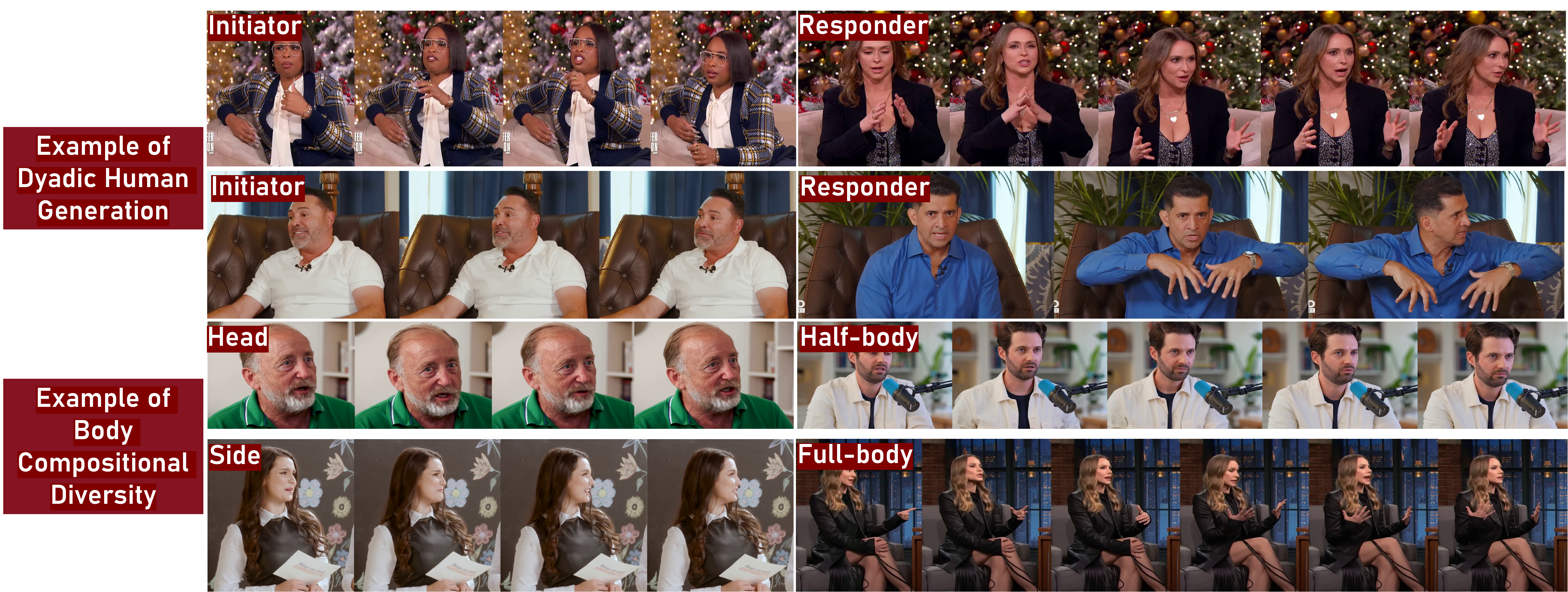}
    \caption{\label{fig:dataset}
    {\textbf{Examples of dyadic dialogue and body composition in SpeakerVid-5M.} The top rows illustrate a typical dyadic human generation sample (initiator and responder). The bottom rows demonstrate the variety of body compositions annotated in our dataset, including close-up headshots, half-body, and full-body views, which are critical for controllable generation.} 
    }
\end{figure*}

\subsection{Dataset Statistical Comparison.}
As shown in Table \ref{tab:compare}, we present a comparative analysis of our dataset against previous general and human video datasets.
Compared to traditional audio- or text-driven digital human generation, our dataset is the first to extend this task into audio-visual dyadic interactive scenarios, featuring complete audio-visual pairs of both questions and responses. This provides rich, high-quality training data for end-to-end audio-visual dyadic interactive virtual human generation. Moreover, our dataset represents the largest collection of single-speaker audio-video pairs to date. Each clip features well-aligned audio and video, capturing a clear and uninterrupted instance of speech. 

Figure.~\ref{fig:dataset_stats} shows statistical analysis of SpeakerVid-5M dataset. The blur score distribution, resolution and DOVER quality distribution demonstrate the high quality of our proposed dataset. Video topic and year distribution and the caption words distribution shows that the dataset exhibits diversity in monologue or conversation content.

Key features of our dataset include:
\textbf{(1) Large-scale data}.
The dataset comprises 5.2M single-speaker audio-video clips totaling 8.7K hours, and 770K two-person conversational audiovisual pairs totaling 1.8K hours.
\textbf{(2) High resolution}.
$93\%$ of the videos are in 1080P or higher, and $98\%$ exceed 720P, ensuring high visual fidelity.
\textbf{(3) Rich annotations}.
Each clip is accompanied by fine-grained structured textual annotations, ASR transcriptions, human pose keypoint sequences, motion magnitude scores, and motion blur scores, enabling detailed modeling and analysis.
\textbf{(4) Diverse data formats}.
The dataset includes single-speaker monologues, two-person dialogues, multi-turn conversational dialogues, and listening-human scenarios,  supporting a wide range of human-centric generation tasks.
\textbf{(5) High-quality audio-visual data:}
Multiple filtering strategies across both audio and video modalities ensure that all clips exhibit clean, well-synchronized audiovisual pairs.
\textbf{(6) Tiered dataset design:}
The dataset is organized into a large-scale pretraining subset and a curated high-quality subset ($1.3$K hours), based on multiple quality indicators, facilitating research under varying levels of training resources and computational constraints.
\textbf{(7) Body compositional diversity}: Data instance is captured with labels spanning full-body, half-body, head, and side-view profiles, enabling fine-grained control over framing analysis.


\subsection{SpeakerVid-5M Dialogue Branch}
To support the training of audio-visual dyadic interactive virtual human generation, we introduce a two-person dialogue branch within the dataset, where each sample consists of two audio-visual pairs, one serving as the input and the other as the target response. Unlike condition-controlled generation tasks such as talking head, or monadic generation tasks such as learning-to-listen, dyadic generation requires the model to generate both audio and video responses based on a comprehensive understanding of the input multi-modal content, enabling interaction with the environment in both listening and behaving.
This setting goes beyond conventional modality-aligned generation, demanding stronger comprehension and reasoning capabilities from generative models. 
Our two-person dialogue branch is primarily sourced from real-world conversational scenarios such as interviews, podcasts, news segments, educational videos, and debates. It consists of 770K clip pairs (totaling 1.8K hours) and includes 16K unique speaker IDs. The dialogue topics span a wide range of categories(Fig.~\ref{fig:c}~\ref{fig:d}, including entertainment, people and blogs, comedy, news and politics, education, and science. 
To ensure dialogue coherence, we only select temporally continuous clips and extract data exclusively from original videos in which the two main speakers account for over $80\%$ of the total speaking time. 

\subsection{SpeakerVid-5M Single Branch}
The single-speaker video branch consists of 5.2M clips with 83K unique speaker IDs, totaling 8.7K hours. It covers a diverse range of camera framings and angles, including full-body, half-body, frontal, and side-view shots. Each clip is annotated with skeletal keypoint sequences, structured textual descriptions, and automatic speech recognition (ASR) transcriptions, making it well-suited for a wide array of conditioned generation tasks, such as audio-skeleton driven portrait animation, lip synchronization, and talking head generation.
In terms of scale, this branch represents the largest speaker-specific dataset to date, with a volume comparable to that of more general-purpose video datasets.

\subsection{SpeakerVid-5M Listening Branch}
Considering the real-world applications of digital humans, models are expected not only to generate meaningful responses but also to exhibit appropriate listening behaviors. To support this capability, we specifically collect two types of data representing listening states:
\textbf{(1) Co-present listening dialogue} (\textit{e.g.}, live interviews, news commentary, seminars, podcasts).
For naturally occurring two-person on-screen dialogues, we filter clips based on audio segmentation results and SyncNet confidence scores. Given a clip with two individuals $A$ and $B$ (before cropping), if only one speaker is active and the SyncNet scores between the two differ larger than a predefined threshold, if $A$ is the lower one, $A$ is determined to be in a \textit{listening state}. 
\textbf{(2) Non-co-present listening dialogue}.
For dialogue scenarios where speakers are not simultaneously visible, we similarly rely on SyncNet scores. A person (suppose as $A$) is classified as being in a \textit{listening state} if the following conditions are met: the ASR result is valid, the transcription confidence is above a given threshold, and the detected SyncNet score for that person's video is below a pre-defined threshold. 
In both cases, the resulting listening pair is composed of the speaker's audio track and the listener's silent video track.


\subsection{SpeakVid-5M Multi-turn Branch}
To enable multi-turn dialogue capabilities in dyadic interactive scenarios, we select multiple clips extracted from the same original video, assigning them with sequential index, and preserve their temporal order. We specifically collect and organize two types of data for multi-turn dialogue. For a given pair of two-person dialogue clips, we define the dialogue start timestamp as $x$ and the maximum history temporal length as $T$. All clips occurring within the interval $[x-T,x]$ are considered as preceding turns of the current dialogue:
\textbf{(1) Contextual multi-turn dialogue}.
The corresponding ASR transcriptions of these preceding turns are aggregated to form the multi-turn dialogue context for the current response prediction.
\textbf{(2) Sequential multi-turn dialogue}.
We exam the temporal gap between consecutive clips. If the gap between two adjacent clips from preceding turns is less than a predefined threshold $\delta{t}$, we consider them part of the same continuous conversation. The audio-visual data from all such clips are then utilized for predicting the current response. (We discard clips shorter than 3s during pre-processing.)
To increase the amount of multi-turn data, we relax the filtering criteria for clips that fall between two valid adjacent dialogue segments. Relaxed criterion facilitates the construction of longer and more natural conversational sequences, enabling the training of dialogue models capable of coherent, multi-turn interactions.

\subsection{Data Stratification}
Beyond the four categorical divisions, we also graded the data by quality. We created a high-quality supervised fine-tuning (SFT) subset (571K clips, $1368$ hours) by selecting samples with a hand motion blur score above $0.5$ and face blur score above $0.7$, a DOVER score above $0.6$. We also constrain the motion score to above 2 and filtering the asr quality with confidence score above $-1$. The remaining data forms the large-scale pretraining subset ($7375$ hours).


\section{Autoregressive Talking Human Generation}
\label{sec:method}

\begin{figure}[h]
    \centering
\includegraphics[width=\linewidth]{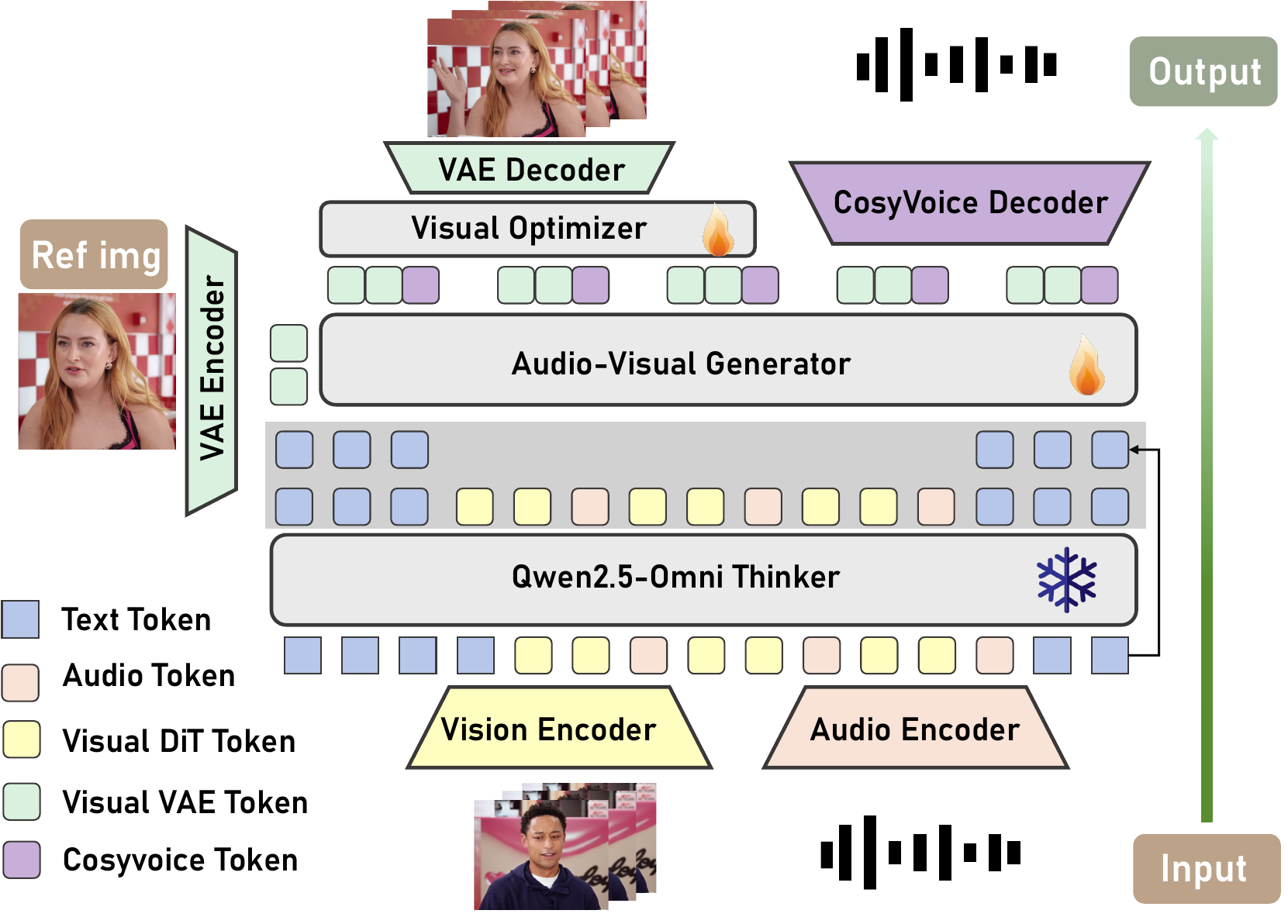}
    \caption{\label{fig:method}\textbf{Our autoregressive audio-visual generation method.} 
    }
\end{figure}

We design a baseline method based on an autoregressive framework tailored for audio-visual dyadic human generation. As illustrated in Fig~\ref{fig:method}, we incorporate Qwen2.5-Omni \cite{xu2025qwen2} to enable multimodal understanding of the input video and audio. Subsequently, a next-chunk prediction autoregressive model is employed to jointly generate audio and video tokens. These tokens are then used as conditioning signals for a shallow diffusion MLP, which produces videos with enhanced detail and realism.
\subsection{Autoregressive generation of video and audio}
Following Qwen2.5-Omni, we feed both the hidden states produced by the Qwen2.5-Omni thinker and the embeddings of the original audio-video inputs into the autoregressive generation head. 
Besides, we employ open-source 3D variational autoencoder (VAE) with a temporal stride of 4 and a spatial stride of 8 to encode the video frames to the latent space \cite{lin2024open}. These latents are then divided into patches and encoded as token embeddings. For audio, we borrow audio tokenizer from CosyVoice2 \cite{du2024cosyvoice} to encode raw audio into discrete tokens. 
We consider a latent map from the 3D-VAE and its corresponding audio tokens to constitute a single chunk. During attention computation, both audio and video tokens attend to all preceding tokens and the tokens within their current chunk. Video tokens are augmented with a combined 1D temporal and 2D spatial positional encoding. Audio tokens utilize a dual-level 1D positional encoding scheme, which encodes both the token's position within its chunk and the chunk's position in the overall sequence. The reference image is also encoded by 3D VAE. Reference image tokens are appended after the input audio and video and before visual generation tokens. To mitigate error accumulation in autoregressive generation, we introduce random noise~\cite{valevski2024diffusion} to the visual tokens during training, which encourages the model to learn more robust representations and leads to improved generation quality.
\subsection{Visual Optimization}
Inspired by MAR~\cite{li2024autoregressive}, we also employ diffusion loss and masked-token prediction to enhance visual generation. For tokens in each generated chunk, another spatial transformer from NOVA~\cite{deng2024autoregressive} takes them as input and generate a detailed visual token set-by-set. The simultaneous generation of audio-video tokens via next-chunk prediction, combined with a carefully designed frame-wise audio injection into the spatial transformer, ensures high audio-visual consistency. The visual tokens generated by the spatial transformer are temporally and spatially aligned with the latents encoded by the 3D VAE, each token corresponds to a specific patch of visual latent feature and is used as a condition signal for diffusion MLP, which performs a denoising process to generate refined visual latents that can be directly decoded by the 3D VAE. CosyVoice flow matching vocoder are used to decode the generated audio tokens into audio.
\subsection{Progressive Training Process}
The model training process is divided into three stages: visual pretraining, audio-visual joint training, and high-quality dyadic dialog fine-tuning.
In the \textbf{visual pretraining} stage, we utilize single-speaker data, where the ASR transcription of the target audio and textual captions (including both motion and expression captions) are used as condition signals to generate video. This stage aims to train the model's basic capability in visual content generation.
In the \textbf{joint audio-visual training} stage, after filtering the audio data, we continue to use the ASR transcription and captions of the target audio as inputs, but extend the generation targets to include both video and audio modality. This stage enables the model to learn synchronized audio-visual generation.
In the \textbf{high-quality dyadic dialog fine-tuning} stage, we select premium dialogue audio-video pairs to further fine-tune the model. The goal of this stage is to enhance its multimodal understanding capabilities and its ability to generate coherent, emotionally aligned conversations.
During training, the visual objective is optimized using a diffusion loss, while the audio objective is supervised with a cross-entropy loss for next-chunk prediction.

\section{Experimental Results and Analysis}
\label{sec:expriment}
\begin{figure*}[t]
    \centering
\includegraphics[width=\textwidth]{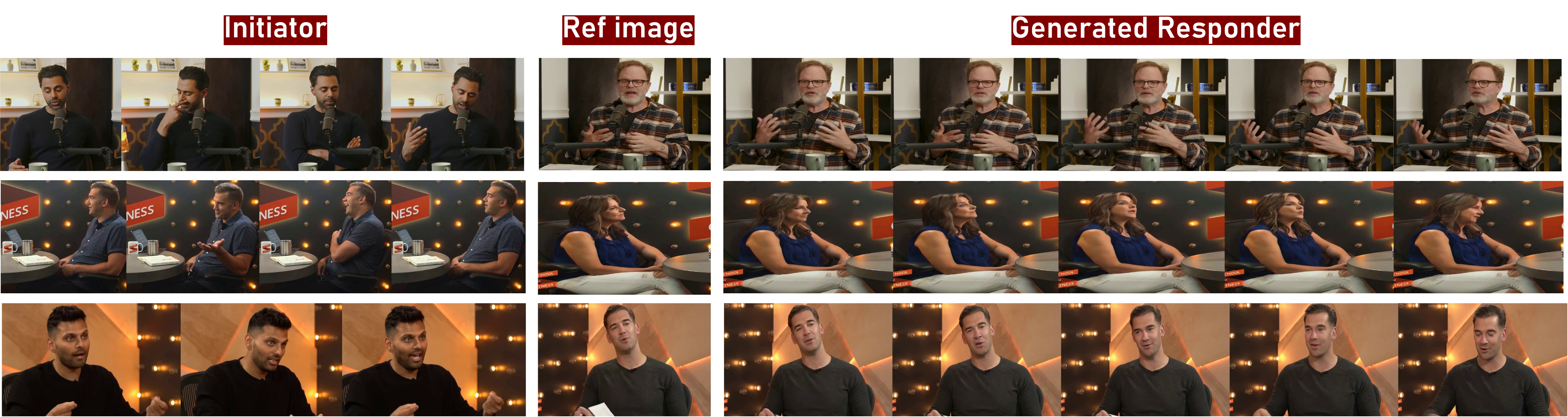}
    \caption{\label{fig:gene}
    {\textbf{Qualitative results of our dyadic generation model.} From left to right, the input video of the initiator, the reference image providing the target identity, and the mode's generated audio-visual response.}
    }
\end{figure*}

\begin{table*}
\begin{center}
\scalebox{0.9}{
\footnotesize
\begin{tabular}{ccccccccccccc}
\toprule
Method & Audio & Spatial & Noise  & FID $\downarrow$ & FVD $\downarrow$ & PSNR $\uparrow$ & SSIM $\uparrow$ & ArcFace $\uparrow$ & $\text{CLIP}_{dialog}$ $\uparrow$ & $\text{Sync}_{conf}$ $\uparrow$ & $\text{FID}_{Emotion}$ $\downarrow$ & SIM-o $\uparrow$\\
\midrule
Conditioned & & & & 56.82 & 55.06 & 15.26 & 0.62 & 0.638 & -- & --&3.45 & -- \\ 
Conditioned & \checkmark & & & 57.03 & 55.16 & 15.31 & 0.62 & 0.630 & -- & 2.063 & 3.45 & 0.65\\ 
Conditioned & \checkmark & \checkmark & & 38.53 & 34.64 & 16.79 & 0.64 & 0.732 & -- & 2.459 & 3.36 & 0.64 \\ 
Conditioned & \checkmark & \checkmark & \checkmark & 34.72 & 30.43 & 17.39 & 0.65 & 0.758 & -- & 2.655 & 3.23& 0.65 \\ 
\midrule
Dyadic & & & & 49.97 & 47.23 & 15.74 & 0.62 & 0.637 & -- & -- & 3.48 & --\\ 
Dyadic &\checkmark &   & &  49.86 & 36.90 & 15.63 & 0.62 & 0.635 & 0.642 & 2.239 & 3.43 & 0.64\\ 
Dyadic &\checkmark & \checkmark &  & 35.67 & 31.28 & 17.44 & 0.65 & 0.749 & 0.643 & 2.541 & 3.33 & 0.65\\ 
Dyadic &\checkmark & \checkmark & \checkmark & \textbf{32.35}  & \textbf{28.82} & \textbf{17.55} &\textbf{0.66} & \textbf{0.772} & \textbf{0.643} & \textbf{2.698} & \textbf{3.22} & 0.65\\ 
\bottomrule
\end{tabular}
}
\caption{
{\textbf{Quantitative results of our baseline method on VidChatBench benchmark.} Audio means generating audio jointly. Spatial means the set-by-set prediction spatial transformer. Noise means the addition of noise in training the autoregressive audio-visual generator.}}
\label{table:main_res}
\end{center}
\end{table*}

\subsection{The VidChatBench Benchmark.}
The audio-visual dyadic interactive virtual human generation involves audio-visual understanding, response generation, and the synthesis of corresponding audio and video content. It requires models to possess both understanding and generative capabilities, making it significantly more challenging than traditional generation task. To effectively evaluate the quality and appropriateness of the generated content, we constructed the \textbf{VidChatBench} benchmark, consisting of 500 representative input-output pairs with unseen speaker IDs, accompanied by a set of tailored evaluation metrics.
Specifically, we assess the model performance along the following five dimensions:
\textbf{(1) Video Quality}. To assess the visual fidelity of the generated videos compared to the ground truth (GT), we adopt several widely used metrics in the video and audio domains, including FID, FVD, PSNR, and SSIM.
\textbf{(2) Identity Preservation}. The generated video is expected to maintain a consistent identity with the reference image throughout the sequence without temporal degradation. We employ ArcFace to extract facial features frame-by-frame from the video and compute the cosine distance between these features and the reference image. The average distance across all frames is used as the identity preservation score.
\textbf{(3) Dialogue Coherence}. The generated content should exhibit semantic relevance and appropriateness with respect to the input. For each evaluation sample in our test set, we construct $5$ candidate responses of varying quality, generated and ranked by a multimodal large language model~\cite{bai2025qwen2}. A score is then assigned to each candidate according to its rank, using the ordered set $[0.2, 0.4, 0.6, 0.8, 1.0]$. During testing, we obtain the ASR transcription of the generated audio and compute its CLIP~\cite{radford2021learning} distance to each candidate response. The final dialogue coherence score is the score of the closest-matching candidate.
\textbf{(4) Audio-Visual Consistency}. Synchronization between audio and lip movements is critical for the perceived realism of the generated video. We evaluate this alignment using the SyncNet confidence score, where a higher score indicates better synchronization between the generated audio and lip motion.
\textbf{(5) Emotional Alignment}. In human interaction, the generated portrait should exhibit emotionally appropriate and temporally coherent expressions. We utilize Deep3DFaceRecon \cite{deng2019accurate} to extract 64-dimensional expression features from both the generated and GT videos. The FID of these expression features measures the emotional alignment between the generated and GT videos.
\textbf{(6) Speaker Identity Preservation}. Following \cite{chen2024f5}, we compute the timbre similarity (SIM-o) for evaluating the timbre of generated audio. The timbre of our generated audio is controlled by the reference audio of the CosyVoice decoder.

\subsection{Qualitative and Quantitative Results}

Our primary quantitative evaluations and ablation studies are presented in Table \ref{table:main_res}, conducted on VidChatBench. We assess our model under two distinct protocols:
The \textbf{Conditioned} Setting, where generation is conditioned on textual inputs, specifically the ground-truth Automatic Speech Recognition (ASR) transcript and a descriptive caption.
The \textbf{Dyadic} Setting, where the model generates a response directly from the audiovisual input of an initiator, thereby simulating a natural dyadic interaction.

The results reveal several key insights. First, the dyadic setting significantly outperforms the conditioned setting, which we attribute to the richer, fine-grained information preserved in direct audiovisual inputs compared to abstracted textual information. Second, our joint audiovisual generation approach successfully maintains high video quality, demonstrating that incorporating audio as an additional condition does not degrade visual fidelity. Furthermore, our ablation studies validate two crucial components: (1) the spatial transformer, which yields substantial improvements in visual metrics by refining frame-level visual tokens, and (2) the training noise injection strategy, which effectively mitigates error accumulation in the autoregressive process, enhance overall video generation quality.

\section{Conclusion}

We introduce the first large-scale high-quality dataset SpeakerVid-5M, which is designed for audio-visual dyadic interactive virtual human generation task as well as VidChatBench benchmark to evaluate performance of trained models. In addition, we provide a baseline dyadic interactive method trained on SpeakerVid-5M. We categorize SpeakerVid-5M into large-scale pretraining subset and a curated high-quality subsets for SFT. Experiments demonstrate the effectiveness of our proposed dataset and the high-quality Supervised Fine-Tuning (SFT) data.

\section{Limitation and Future Work}
While SpeakerVid-5M represents a significant contribution to the field of interactive virtual humans, we still discuss the current scope of this work and highlight several promising avenues for further development.
First, our baseline model is presented as a preliminary benchmark to validate the SpeakerVid-5M dataset. Its performance was constrained by limited computational resources and does not represent the state-of-the-art. We anticipate that future work using more advanced architectures and large-scale training will unlock the dataset's full potential and yield higher-fidelity results.
Second, the scope of interaction within SpeakerVid-5M is focused on single-person and two-person (dyadic) scenarios. The dataset does not currently capture the complex dynamics of multi-party conversations, such as group turn-taking. Extending the data collection to include these multi-person interactions is valuable for future research.

\maketitlesupplementary

\section{Implementation Details}
During both training and inference, we standardize the frame rate to 8 FPS and resize all video frames to 480×768 resolution. The VAE compresses the input with a temporal stride of 4 and a spatial stride of 8.
We treat each 4×4 feature patch as a token embedding, resulting in 360 tokens per frame in the latent space. The spatial transformer further refines each latent frame into 1440 tokens. For the audio modality, each chunk contains 12 audio tokens. To initiate generation, we introduce learnable special tokens as start-of-generation embeddings for both the audio and visual tokens.
The Qwen2.5-Omni Thinker is kept frozen throughout the training process, while all other components remain trainable, resulting in a total of 0.8 billion trainable parameters. The learning rate is set to 1e-4, with warm-up and decay strategies applied.
The visual pretraining and joint audio-visual training are conducted over 15 days on 128 NVIDIA L40S GPUs, with video clips in 3-7 seconds. The fine-tuning stage is carried out on 32 NVIDIA A800 GPUs over 5 days, using clips ranging from 3 to 14 seconds.

\section{Annotation File Usage}
In this section, we provide detailed explanation of the annotation files in our SpeakerVid-5M dataset to promote the application.
The basic annotation file serves as a central repository for essential metadata pertaining to each clip. This includes:
Source video ID and URL: Enabling direct access to the original video content.
Clip timestamps: The precise start and end times of the clip within the source video.
Human bounding box: Spatial localization of the primary subject within the clip's frames.
Video resolution: The dimensions of the video content.
SyncNet confidence score: A metric assessing the lip-sync quality.
DOVER score: A quality score of overall video clip quality.
Clarity score: An indicator of visual clarity.
Speaker ID: Unique identification based on audio for the speaker present in the clip.
These comprehensive annotations are designed to empower users to directly retrieve the original videos from YouTube and efficiently extract high-quality single-person audiovisual clips using straightforward FFmpeg commands. Furthermore, we provide open-source code that facilitates the intelligent linking of these individual clips, thereby enabling the convenient construction of two-person dialogues and more complex multi-turn dialogue datasets.

\textbf{The \texttt{l\_score} Annotation.}
The \texttt{l\_score} annotation provides quantitative clarity metrics for salient regions within each video, namely the left hand, right hand, and face. For each region, we compute Laplacian-based scores on a frame-by-frame basis. These are provided as both {absolute scores}, which are comparable across the entire dataset, and {relative scores}, normalized within each clip. Furthermore, the frame-wise absolute scores are aggregated to produce a single, holistic {clip-level score}. This hierarchical annotation is designed for two primary use cases: the fine-grained, frame-level scores serve as dynamic conditioning signals for generative models, enabling more precise control over motion fidelity. In contrast, the clip-level scores offer an effective mechanism for dataset curation, allowing researchers to easily filter for high-clarity training samples to enhance model robustness.

\textbf{The \texttt{Caption} Annotation.}
The \texttt{Caption} annotation file provides a rich set of clip-level semantic labels, automatically generated using large multimodal model (qwen2.5 vl). These annotations offer a structured, multi-faceted description of each video's content, enabling detailed analysis and control. The specific labels are as follows:
\begin{itemize}
    \item \textbf{Video Quality:}
    \begin{itemize}
        \item \textbf{Clarity:} A binary label indicating the absence or presence of significant motion blur or artifacts.
        \item \textbf{Camera Dynamics:} Classification of camera motion (e.g., static, panning, zooming) and shot framing (e.g., close-up, medium shot).
    \end{itemize}
    \item \textbf{Subject \& Scene Composition:}
    \begin{itemize}
        \item \textbf{Subject Count:} The number of individuals detected in the frame.
        \item \textbf{Framing View:} A label indicating if the subject is framed in a full-body or upper-body view.
        \item \textbf{Head Pose:} The estimated orientation of the subject's head (e.g., frontal, side view).
        \item \textbf{Scene Entities:} A list of recognized objects and persons present.
    \end{itemize}
    \item \textbf{Behavioral \& Action Details:}
    \begin{itemize}
        \item \textbf{Speech Activity:} A binary label indicating active speech.
        \item \textbf{Motion Status:} A binary label classifying the primary subject as either static or in motion.
        \item \textbf{Movement Intensity:} A categorical rating of the subject's motion intensity (e.g., low, medium, high).
        \item \textbf{Holistic Action Summary:} A high-level textual description of the subject's overall actions.
        \item \textbf{Fine-grained Action Description:} A detailed textual account of specific, nuanced actions performed by the subject.
        \item \textbf{Facial Expression Summary:} A textual description of the subject's primary facial expression.
    \end{itemize}
\end{itemize}
Collectively, these annotations furnish a structured representation of the video content. This enables both fine-grained conditioning for generative tasks and provides a robust framework for the objective evaluation of synthesized visual attributes and behaviors.

\textbf{The \texttt{Scene} Annotation.}
The \texttt{Scene} annotation provides the temporal metadata required for managing and assembling video clips from the original source footage. This annotation is generated through a two-stage process. First, an initial scene detection algorithm segments the source videos into coherent shots, recording their start/end timestamps and frame indices. Subsequently, these segments are further partitioned to enforce a maximum duration of 14 seconds per clip. The final annotation file catalogs each processed clip with a unique identifier and its precise temporal boundaries (start and end times). This structured metadata is crucial for downstream applications, enabling the seamless concatenation of clips for long-form video synthesis and providing a flexible framework for tasks involving multi-turn dialogue or extended narrative generation.

\textbf{The \texttt{Speaker} Annotation.}
\texttt{Speaker} annotation is performed on the original audio of video recordings to provide detailed speaker diarization results. This comprehensive annotation process involves several key components.
Initially, raw diarization results are generated, capturing the precise start and end times for each detected speaker turn, along with their respective assigned speaker IDs. These results provide a foundational temporal map of all spoken segments.
For applications focused on dyadic dialogue scenarios, the raw results undergo a filtering process. This step specifically identifies and prioritizes the two primary speakers (labeled as Speaker A and Speaker B). A critical criterion for this filtering is that these two speakers must collectively account for at least 80\% of the total speech duration within the video. Furthermore, in the context of dyadic data filtering, only the conversational exchanges occurring exclusively between Speaker A and Speaker B are retained. This ensures the selection of highly relevant dialogue for subsequent model training and analysis.
The final stage yields a cleaned and refined list of speaker IDs and their corresponding start and end times for each segment. This provides clear and accurate speaker attribution of the video, ensuring high-quality data for downstream tasks.

\textbf{The \texttt{ASR} Annotation.}
\texttt{ASR} annotation is performed on unified single-person audiovisual clips, which are meticulously derived from praevia video and audio processing. For example, in dual-person co-present scenes, the segmentation process spatially divides the video into two distinct regions, each corresponding to an individual. Concurrently, the associated audio is temporally segmented to align with the speech of each respective person.
To achieve these single-person audiovisual clips—where the video prominently features only one person and the audio contains solely that person's speech—we integrate YOLO for visual analysis with speaker diarization for audio segmentation. Upon obtaining these refined clips, automatic speech recognition is then applied to their corresponding audio segments. The ASR output encompasses several key components.
Transcribed speech text: The verbatim textual representation of the spoken content.
Confidence score: A metric indicating the reliability and accuracy of the transcription.
Speech compression ratio: The ratio between the original speech duration and its compressed form, relevant for efficiency analysis.
No-speech probability: The likelihood that a given audio segment contains no discernible speech.
Language information: Identification of the language spoken within the clip.

\section{Visualization of Pretrain and Finetune Model}
\begin{figure*}[t]
    \centering
\includegraphics[width=\textwidth]{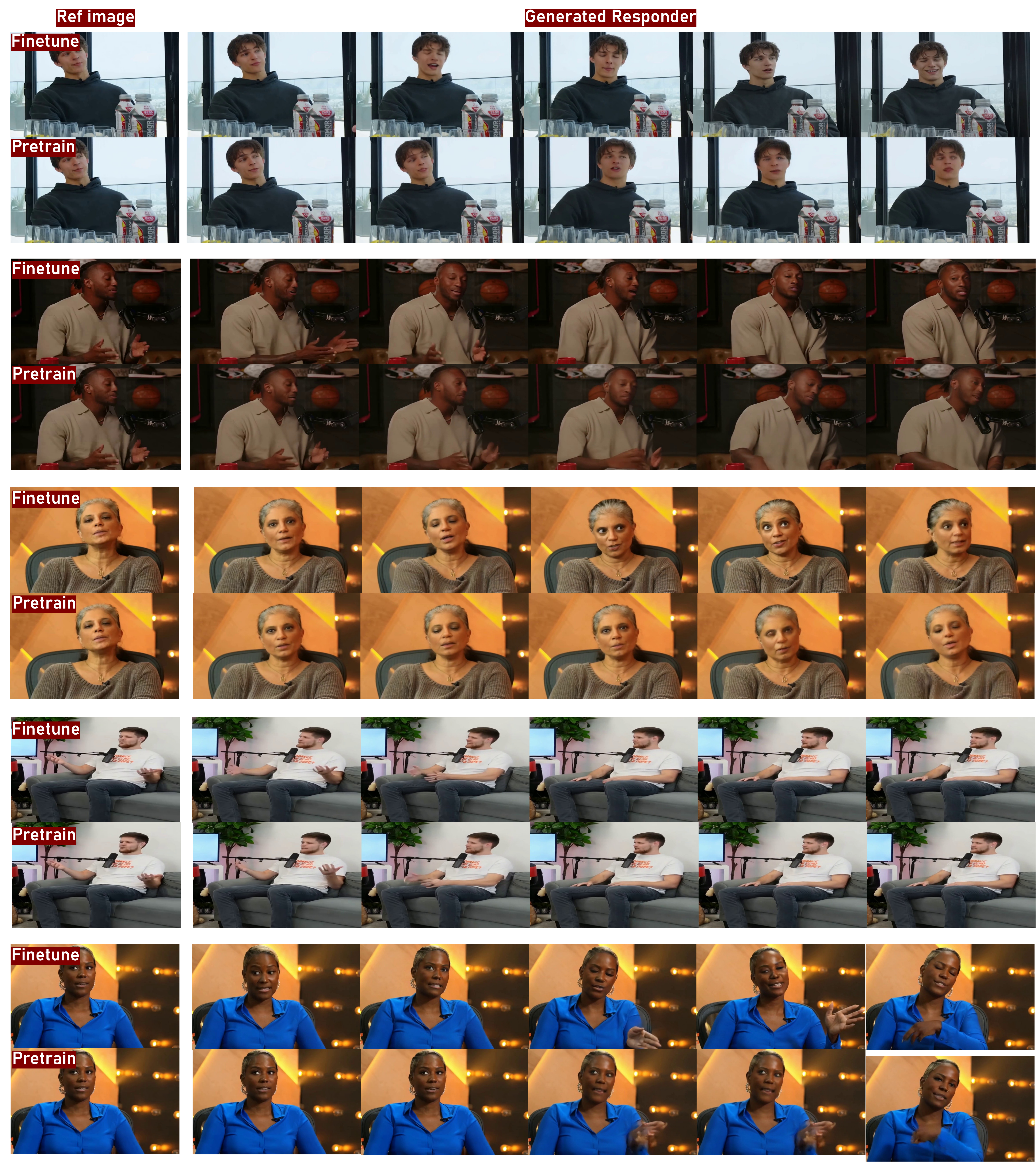}
\caption{\label{fig:pretrain}
{\textbf{Impact of finetuning on generation quality.} A comparison between the model after pretraining phase and after finetuning on our high-quality subset.    }}
\end{figure*}

As illustrated in Figure \ref{fig:pretrain}, the model resulting from our initial pre-training phase exhibits pronounced motion blur, particularly in the face and hands. These artifacts are significantly exacerbated during rapid movements, degrading perceptual quality. To address this, we curated a high-quality subset by filtering the training data based on facial and hand clarity. The figure demonstrates that fine-tuning on this refined data yields substantial improvements in visual fidelity, markedly reducing blur and enhancing detail in regions critical for human perception.
To enable this data refinement, we compute blur scores for both the face and hands in each frame using a Laplacian-based method. Recognizing their broader utility, we release these scores as part of our public dataset annotations. The value of such explicit quality signals is substantiated by prior work; for instance, CyberHost demonstrated that conditioning a model on hand blur scores can significantly enhance the clarity and fidelity of synthesized hand motions.

\begin{figure*}[t]
    \centering
\includegraphics[width=\textwidth]{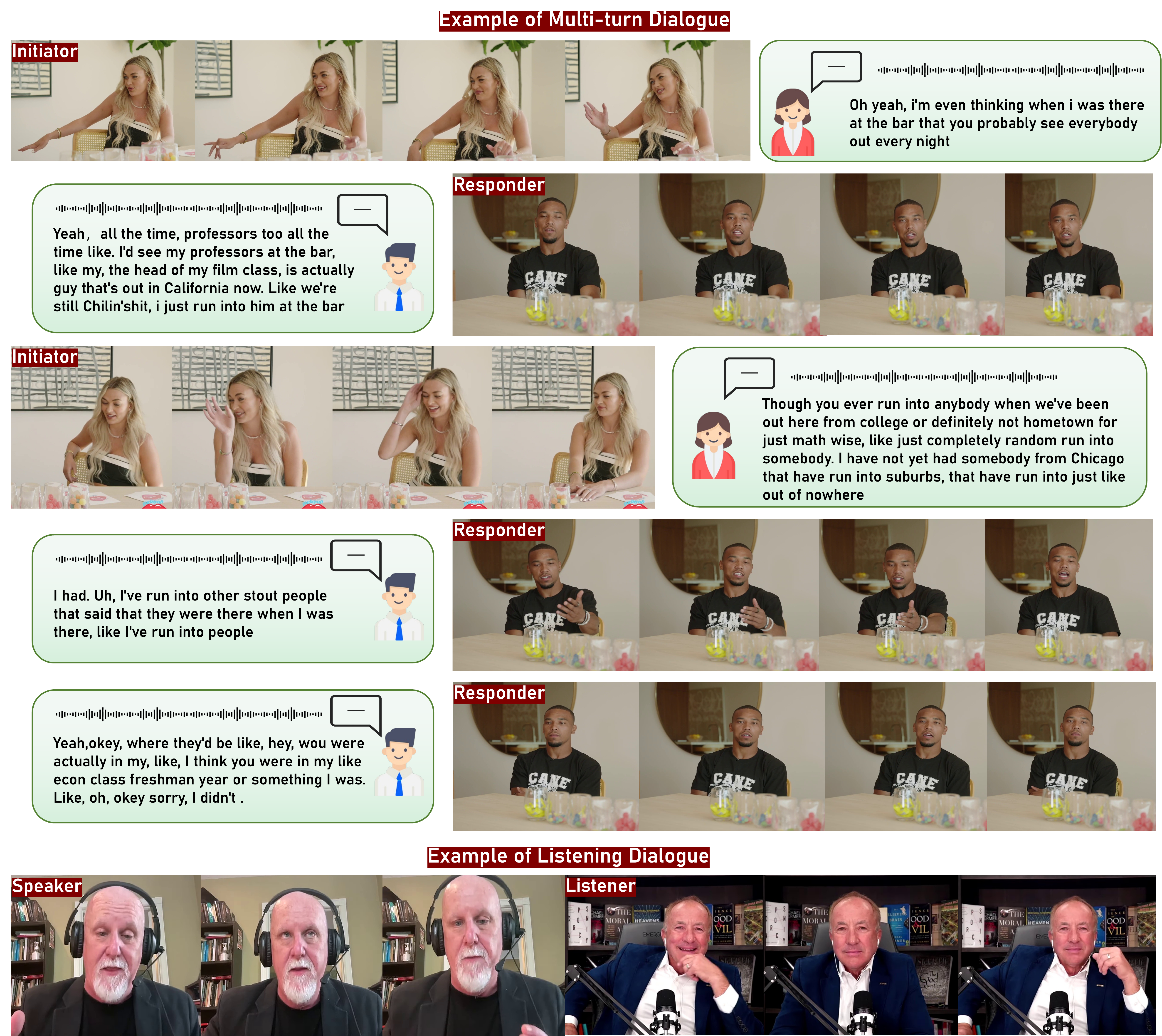}
\caption{\label{fig:multiturn}
{\textbf{Visualization of multi-turn dialogue and listening scenarios.} Top (Multi-turn Dialogue): We showcase a sequence of conversational turns between an initiator and a responder. By preserving temporal context, our dataset facilitates the training of models capable of coherent, long-form conversations. Bottom (Listening Case): A speaker is paired with a non-speaking listener.}
    }
\end{figure*}

\begin{figure*}[t]
    \centering
\includegraphics[width=\textwidth]{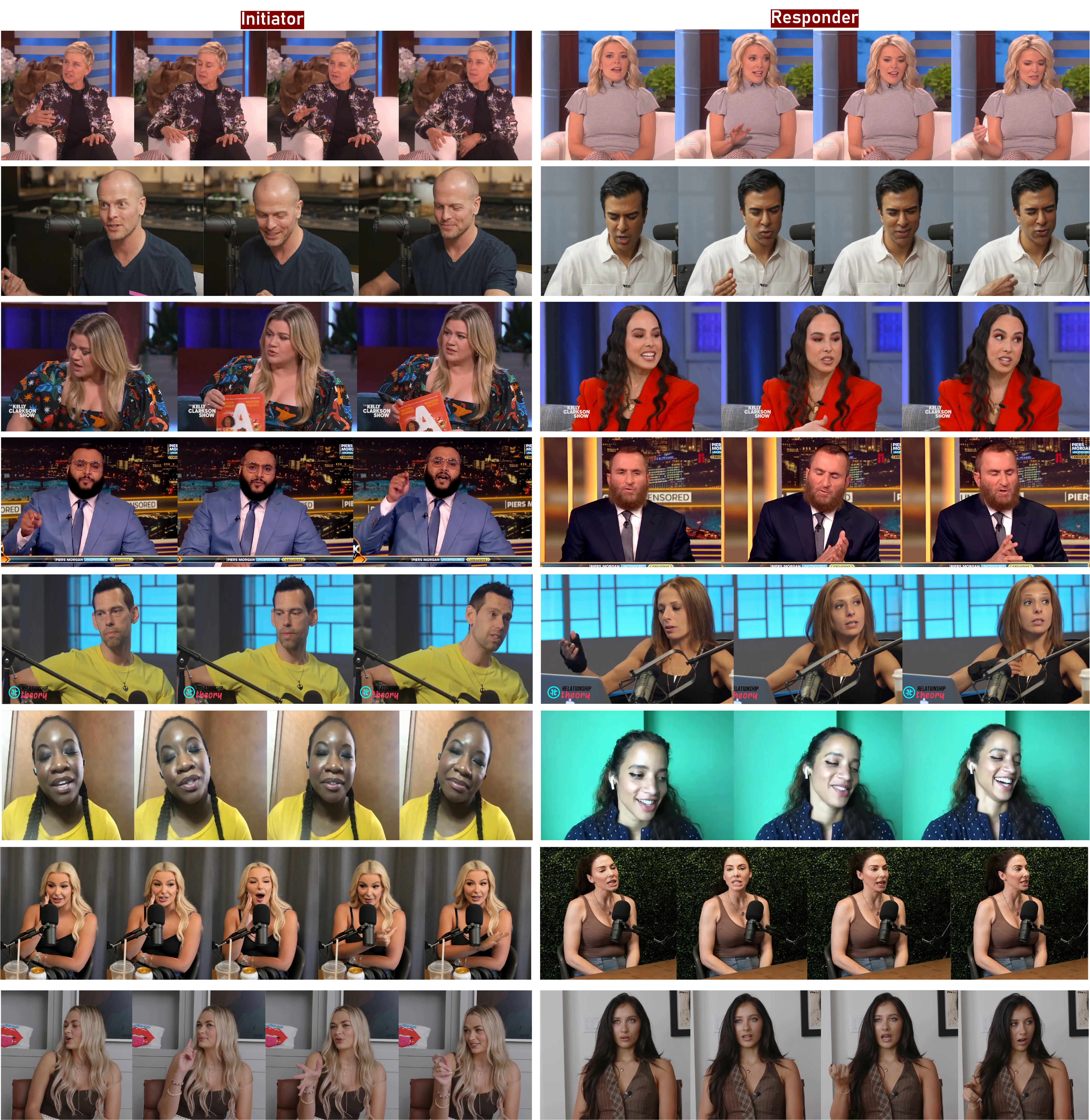}
\caption{\label{fig:more_dataset}
{\textbf{Diverse examples of dyadic pairs from SpeakerVid-5M.} This figure highlights the diversity of subjects, environments, and interaction styles captured in our dataset.}
    }
\vspace{5mm}
\end{figure*}

\section{Prompt Used in Annotation}
Tables \ref{tab:video_annotation_prompt} and \ref{tab:motion_annotation_prompt} present the caption prompts used for Qwen-VL-2.5, which serve as effective guidance for generating high-quality responses during captioning.

\begin{table*}[h!]
\caption{Prompt Used in Video Annotation}
\label{tab:video_annotation_prompt}
\begin{tabularx}{\textwidth}{@{}>{\raggedright\arraybackslash}X@{}}
\toprule

\textbf{\# Your Role:} Video Annotation Expert \\
\\
\textbf{\#\# Objective:}  
As a professional annotator, your task is to evaluate and label video content based on clarity, camera motion, human presence and activity, and semantic understanding. Your annotations must be structured and follow a predefined JSON format. \\
\\
\textbf{\#\# Annotation Guidelines}  
For each input video, assess and annotate the following attributes:  
\begin{enumerate}
    \item \textbf{Visual Clarity:} Assess whether the primary subject in the video is clearly visible and distinguishable. Options: \texttt{"Yes"} or \texttt{"No"}.
    
    \item \textbf{Camera Motion:} Identify the type(s) of camera movement observed. Choose one or more from the following: \texttt{["Static Camera", "Dynamic Shooting", "Still Shot", "Left-to-Right Pan", "Right-to-Left Pan", "Zoom In", "Zoom Out"]}.
    
    \item \textbf{Human Motion Presence:} Determine whether the person in the video exhibits obvious and recognizable physical movement. Options: \texttt{"Yes"} or \texttt{"No"}.
    
    \item \textbf{Motion Intensity Level:} Rate the level of physical activity on a scale from 1 (very minimal motion) to 5 (extremely intense movement). Aim to differentiate clearly across the full range rather than concentrating ratings in the middle.
    
    \item \textbf{Entity List:} List all visually prominent entities in the video in descending order of saliency, e.g., \texttt{["man", "teddy bear", "river", "traffic sign", "apple"]}.
    
    \item \textbf{Speech Presence:} Identify whether the person is clearly speaking in the video. Options: \texttt{"Yes"} or \texttt{"No"}.
    
    \item \textbf{Observed Actions:} List the specific actions shown in the video, such as \texttt{["running", "dancing", "eating", "talking", "singing", "presenting"]}.
    
    \item \textbf{Number of People:} Count the number of distinct people visible in the video.
    
    \item \textbf{Upper-Body Only:} Indicate whether only the upper half of the person is visible (if the legs or lower body appear, select "No"). Options: \texttt{"Yes"} or \texttt{"No"}.
    
    \item \textbf{Facing Direction:} Specify the subject’s orientation relative to the camera. Options: \texttt{"Front"}, \texttt{"Side"}, or \texttt{"Back"}.
\end{enumerate}

\\
\textbf{\#\# Output Format:}  
Return the annotation results in the following JSON structure:
\begin{enumerate}
    \item \textbf{Clarity:} \texttt{"Yes"}
    \item \textbf{Camera Motion:} \texttt{["Zoom In", "Left-to-Right Pan"]}
    \item \textbf{Human Motion Presence:} \texttt{"Yes"}
    \item \textbf{Motion Intensity Level:} \texttt{[3]}
    \item \textbf{Entity List:} \texttt{["man", "dog", "apple"]}
    \item \textbf{Speech Presence:} \texttt{"Yes"}
    \item \textbf{Observed Actions:} \texttt{["talking"]}
    \item \textbf{Number of People:} 1
    \item \textbf{Upper-Body Only:} \texttt{"Yes"}
    \item \textbf{Facing Direction:} \texttt{"Side"}
\end{enumerate}
\\
\textbf{Note:}  
Please ignore any subtitles or on-screen text when performing the annotation. Focus solely on the video’s visual and auditory content. \\

\bottomrule
\end{tabularx}
\end{table*}

\begin{table*}[h!]
\caption{Prompt Used in Video Motion Intensity Level}
\label{tab:motion_annotation_prompt}
\begin{tabularx}{\textwidth}{@{}>{\raggedright\arraybackslash}X@{}}
\toprule
\textbf{\# Your Role:} Human Motion Expert \\
\\
\textbf{\#\# Annotation Guidelines:}
As a human motion expert, assess the intensity of movement in the video by focusing on the amplitude and frequency of body movements:
1. Movement Amplitude: Whether the person frequently makes large body movements like arm swings, body turns, etc.
2. Movement Frequency: Whether the person repeats similar movements frequently, such as nodding their head, gesturing, etc. Score range from 1 to 5.
\begin{enumerate}
    \item The person is nearly stationary, with only minimal head or small hand movements.
    \item The person makes occasional small gestures or minor body adjustments.
    \item The person uses moderate gestures with occasional body adjustments, such as slight forward leans.
    \item The person uses frequent gestures, with larger body movements and more frequent body shifts.
    \item The person’s movements are frequent and large, with extensive use of hand gestures and body shifts, indicating high intensity.
\end{enumerate}
\\
\bottomrule
\\
\textbf{\# Your Role:} Audience Member \\
\\
\textbf{\#\# Annotation Guidelines:}
As an audience member, assess the movement intensity of the speaker by focusing on their emotional expression and the resulting body movements:
1. Emotional Fluctuations: Whether the speaker shows large emotional fluctuations in their speech, and whether it is accompanied by body language.
2. Audience Reactions: Whether the speaker adjusts their body in response to audience reactions (e.g., applause, nodding). Score range from 1 to 5.
\begin{enumerate}
    \item The speaker speaks in a calm, even tone with minimal body movement.
    \item The speaker has slight emotional fluctuations, with occasional small gestures or head movements.
    \item The speaker shows moderate emotional fluctuations and occasional body movements like hand gestures or body shifts.
    \item The speaker shows significant emotional fluctuations, with frequent body movements, gestures, and emotional intensity.
    \item The speaker displays intense emotional fluctuations, with frequent and large gestures and body movements, indicating high intensity.
\end{enumerate}
\\
\bottomrule

\\
\textbf{\# Your Role:} Labeling Expert \\
\\
\textbf{\#\# Annotation Guidelines:}
As a labeling expert, assess the intensity of movement by considering the body language and interaction frequency:
1. Gesture Usage: Whether the person frequently uses hand gestures or body movements to emphasize their speech.
2. Interaction Frequency: Whether the person interacts frequently with others, especially through body language responses (e.g., nodding, smiling). Score range from 1 to 5.
\begin{enumerate}
    \item The person makes little to no gestures or body movements and interacts minimally.
    \item The person occasionally uses gestures or makes small body movements, with limited interaction.
    \item The person frequently uses gestures and makes moderate body adjustments, with moderate interaction with others.
    \item The person uses hand gestures frequently, with larger body movements and frequent interaction.
    \item The person has frequent and strong body movements, with highly frequent interactions and large gestures, indicating high intensity.
\end{enumerate}
\\
\bottomrule

\end{tabularx}
\end{table*}

\section{Failed Case and Analysis}
(1) Fine-grained Hand Dynamics: Prevailing talking-head datasets are predominantly constrained to rudimentary hand motions, such as resting, crossing, or waving. Consequently, the synthesis of more complex and expressive actions, including intricate gesturing or human-object interactions, remains a formidable challenge in digital human generation. This difficulty is compounded by the fact that textual descriptions typically lack the requisite granularity to supervise these fine motor skills, thereby hindering a model's ability to learn rich hand dynamics. To address this limitation, we provide comprehensive annotations for each video clip, encompassing hand keypoints, bounding boxes, and a novel hand-blur score. These annotations furnish precise supervisory signals, enabling more effective control and rigorous evaluation of synthesized hand motion quality.

(2) Synthesis of Occluded or Out-of-Frame Body Parts: A core challenge in this domain is the synthesis of body parts that are occluded or absent in the reference image, for instance, rendering a newly raised arm or animating eyes from a closed to an open state. This task necessitates models endowed with robust generative priors and a comprehensive understanding of human anatomy and plausible motion. The challenge is often exacerbated by data bias; many existing datasets are heavily skewed towards front-facing, tightly-cropped portraits, which lack diverse viewpoints and full-body context. To counteract this, our dataset incorporates a wide array of perspectives, including full-body, upper-body, and profile views. This provides a more balanced distribution of human body configurations, thereby empowering models to generalize beyond the visible input and mitigate in-painting artifacts.

(3) Temporal Coherence in Long-Form Video Generation: Generating temporally coherent, long-form video presents two primary obstacles. First, the scarcity of high-quality, long-duration training data fundamentally limits model performance. Our dataset addresses this by providing a large corpus of high-fidelity clips (5–14 seconds) that are specifically designed for seamless concatenation into extended sequences (\textit{e.g.}, 20 seconds to 3 minutes). Second, autoregressive models are prone to error accumulation, which degrades generation quality over time. To mitigate this, we introduce a simple yet effective noise injection strategy during training. This technique enhances the model's temporal stability, ensuring consistent quality over extended generative horizons.
\begin{figure*}[t]
    \centering
\includegraphics[width=\textwidth]{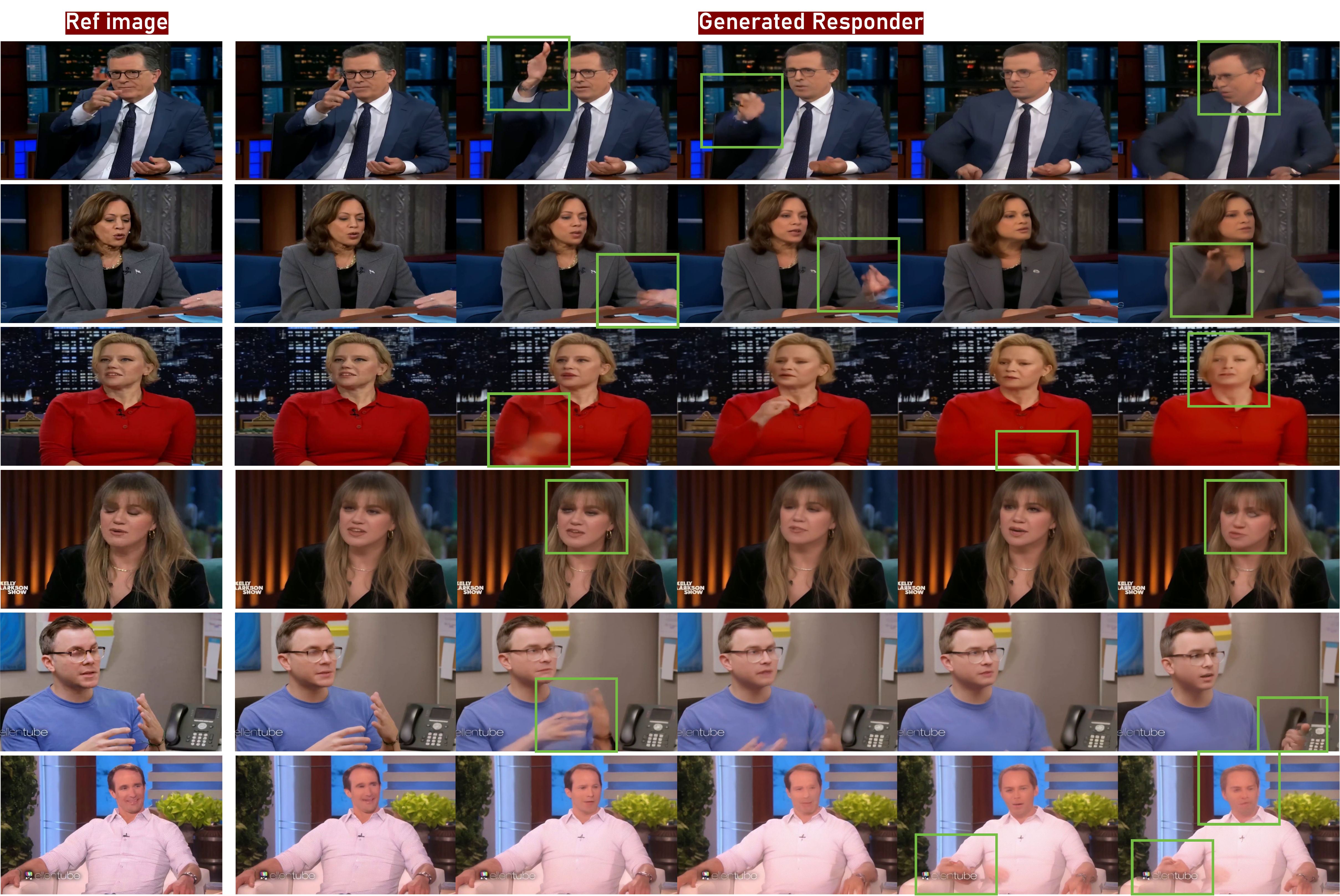}
\caption{\label{fig:failed_case}
{\textbf{Analysis of failure cases.} These include generating facial artifacts or unnatural distortions and struggling with severe motion blur during rapid head or hand movements.
}
    }
\end{figure*}

{
    \small
    \bibliographystyle{ieeenat_fullname}
    \bibliography{main}
}
\end{document}